\pdfoutput=1
\documentclass[11pt, a4paper]{book}
\usepackage{svn-multi}
\svnid{$Id$}
\renewcommand{\thepage}{\roman{page}}
\usepackage{amsthm}
\usepackage{adjustbox}
\theoremstyle{definition}
\newtheorem{definition}{Definition}[section]
\newtheorem{hyp}{Hypothesis}
\theoremstyle{remark}

\usepackage[square]{natbib}
\usepackage{har2nat}
\usepackage[hyperindex=true,
			bookmarks=true,
            pdftitle={}, pdfauthor={},
            colorlinks=false,
            pdfborder=0,
            pagebackref=false,
            citecolor=blue,
            plainpages=false,
            pdfpagelabels,
            pagebackref=true,
            hyperfootnotes=false]{hyperref}
\usepackage{amsmath}
\usepackage{enumerate}
\usepackage[all]{hypcap}
\usepackage[palatino]{anuthesis}
\usepackage{afterpage}
\usepackage{graphicx}
\usepackage{thesis}
\usepackage[normalem]{ulem}
\usepackage[table]{xcolor}
\usepackage{makeidx}
\usepackage{cleveref}
\usepackage[centerlast]{caption2}
\usepackage{float}
\usepackage[T1]{fontenc}
\usepackage[scaled]{beramono}
\usepackage{multirow}
\usepackage{algorithm}
\usepackage{algpseudocode}
\renewcommand*{\backref}[1]{}
\renewcommand*{\backrefalt}[4]{
  \ifcase #1 %
  \or
    (cited on page #2)%
  \else
    (cited on pages #2)%
  \fi
}

\usepackage{booktabs}
\usepackage{relsize}
\usepackage{xspace}
\usepackage{subfigure}
\usepackage{listings}
\definecolor{tableheadcolor}{rgb}{0.8,0.8,1.0}
\definecolor{tablealtcolor}{rgb}{0.9,0.9,0.95}

\definecolor{todocolor}{rgb}{0.8,0.8,1.0}
\definecolor{fixcolor}{rgb}{1,0.8,0.8}
\definecolor{commentcolor}{rgb}{0.8,1.0,0.8}

\usepackage[color=todocolor, colorinlistoftodos]{todonotes}

\newcommand{\textjava}[1]{{\lstset{basicstyle=\ttfamily}\lstinline@#1@}}
\newcommand{\textjavafn}[1]{{\lstset{basicstyle=\footnotesize\ttfamily}\lstinline@#1@}}
\usepackage{setspace}
\usepackage{ifthen}

\long\def\sfootnote[#1]#2{\begingroup%
\def\thefootnote{\fnsymbol{footnote}}\footnote[#1]{#2}\endgroup}
\newcommand{\doi}[1]{\href{http://dx.doi.org/#1}{\nolinkurl{doi:#1}}}
\newcommand{\ignore}[1]{}

\title{Representation Learning for Weakly Supervised Relation Extraction}
\author{Zhuang Li}
\date{\today}

\renewcommand{\thepage}{\roman{page}}

\makeindex
\DeclareUnicodeCharacter{2212}{-}
\begin{document}
\pagestyle{empty}
\thispagestyle{empty}

\begin{titlepage}
  \enlargethispage{2cm}
  \begin{center}
    \makeatletter
    \Huge\textbf{\@title} \\[.4cm]
    \Huge\textbf{\thesisqualifier} \\[2.5cm]
    \huge\textbf{\@author} \\[9cm]
    \makeatother
    \LARGE Master Of Computing(Advanced) \\
    The Australian National University \\[2cm]
    October 2015
  \end{center}
\end{titlepage}

\input{frontmatter}

\cleardoublepage
\pagestyle{empty}
\vspace*{7cm}
\begin{center}
To my parents, for all their love and blessings.
\end{center}

\cleardoublepage
\pagestyle{empty}
\chapter*{Acknowledgments}
\addcontentsline{toc}{chapter}{Acknowledgments}
I would like to express my special appreciation to my supervisor, Dr. Lizhen Qu who supervised me patiently when I was in trouble and confusion. My sincere thanks would also go to Dr. Gabriela for her expert advice, Convenor John Slaney for the consultant of this project. I thank my friend, Qiongkai Xu, for his proofreading within limited time frame. Last but not the least, I would like to thank my parents, Xianlan Yang and Taiping Li, for their selfless supporting through my life.

\cleardoublepage
\pagestyle{headings}
\chapter*{Abstract}
\addcontentsline{toc}{chapter}{Abstract}
\vspace{-1em}
Recent years have seen a rapid development in Information Extraction, as well as its subtask, Relation Extraction.\\
Relation Extraction is able to detect semantic relations between entities in sentences. Currently, many efficient approaches have been applied on relation extraction task. Supervised learning approaches especially have the good performance. However, there are still many difficult challenges. One of the most serious problems is that manually labelled data is difficult to acquire. In most cases, limited data for supervised approaches equals lousy performance. Thus here, under the situation with only limited training data, we focus on how to improve the performance of our supervised baseline system with unsupervised pre-training.\\
Feature is one of the key components in improving the supervised approaches. Traditional approaches usually apply hand-crafted features, which requires expert knowledge and expensive human labor. However, this type of feature might suffer from data sparsity: when training set size is small, the model parameters might be poorly estimated. In this thesis, we present several novel unsupervised pre-training models to learn the distributed text representation features, which are encoded with rich syntactic-semantic patterns of relation expressions. \\
The experiments have demonstrated that this type of features, combine with the traditional hand-crafted features, could improve the performance of logistic classification model for relation extraction, especially on the classification of relations with only minor training instances.


\cleardoublepage
\pagestyle{headings}
\markboth{Contents}{Contents}
\tableofcontents
\listoffigures
\listoftables
\mainmatter

\chapter{Introduction}
There is vast amount of unstructured text data on the World Wide Web, sources of which are like email, blogs, website documents and etc. However this kind of unstructured data is usually difficult for machine or human to analysis. There is a common way to solve this problem, Relation Extraction. Relation extraction is a technique that let computers automatically extract relations we are interested in from text. Then the structured data can be stored in the structural database and easily analysed by machine or human.\\
Basically, there are four approaches for relation extraction, supervised learning approaches, unsupervised learning approaches, semi-supervised learning approaches and distant supervision approaches. Supervised approaches usually have a better performance if with enough training data.\\
However, there are two major limitations of this method. 
\begin{enumerate}  
\item The labelled data is resource consuming and difficult to acquire. Usually we need a lot of human and time resources to annotate the corpus. 
\item Hand-crafted feature of text, such as unigrams and bigrams, usually do not perform well when training data size is small. 
\end{enumerate}
For solving these two problems, a technique named \textbf{Text Representation Learning} can be used to acquire a new kind of distributed representation features. This kind of feature is able to help improve the performance of the supervised approaches especially under the situations with only limited data.
\section{Objectives}
The goal of this thesis is to have a thorough study of text representation learning application on relation extraction issues. The specific objectives are:\\
\begin{itemize}
	\item Apply text representation learning techniques and acquire a new feature of distributed representations.
	\item Combine distributed representation feature and traditional hand-crafted feature to improve the performance of baseline relation extraction system.
\end{itemize}
To achieve these objectives, the procedures are as follows:
\begin{enumerate}
	\item Build the baseline relation extraction systems.
	\item Design text representation learning models.
	\item Combine baseline and text representation learning models to implement a new system.
	\item Design a series of experiments with different variations.
	\item Observe and analysis the results.
\end{enumerate}
\section{Contributions}
Major contributions of the work are as follows:
\begin{enumerate}
	\item The novel models based on representation learning techniques outperform state-of-art baseline systems on extracting relations of interests.
	\item We conducted extensive experiments to study the performance of new models in comparison of baseline systems. 
\end{enumerate}
\section{Dissertation Organization}
This thesis is structured as follows. Chapter 1 is the general introduction of this thesis. Chapter 2 introduces basic concepts of relation extraction and its related works. Chapter 3 covers the most common and popular text representation learning techniques along with the preliminaries for text representation learning techniques in this thesis. Chapter 4 describes baseline system using hand-crafted feature, our models for learning text representations, our motivations behind and the architecture of the new system using both hand-crafted features and text representation features. Chapter 5 places details of a series of experiments that we have designed, including the description of dataset, the settings of hyperparameters and the results of the experiments. Chapter 6 concludes the project and proposes the future work.

\chapter{Relation Extraction}
\label{cha:relationextraction}
Recent years have seen an rapid development of automatic knowledge base construction along with information extraction. As a part of Information Extraction, relation extraction is being paid attention as well. This chapter is to introduce relation extraction from a comprehensive perspective. Since relation extraction is a subtask of information extraction(IE), we will give a brief explaination of information extraction first. Then knowledge base, which is an important preliminary concept in relation extraction, will be introduced. In section \ref{sec:conceptofrelationextraction}, we will offer rough ideas about relation extraction. Last section places some current common machine learning techniques and their application on relation extraction tasks.
\section{Information Extraction}
\label{sec:informationextraction}
It is known that there is a vast amount of unstructured text on the Internet that is not machine-readable and difficult to search in. This results in the "grown need for efficient techniques on text analysing and discovering valuable information"\cite{piskorski2013information} with the structure of our interest automatically.
Basically, information extraction is the technique to turn the unstructured data to the structured form. Then in further,the data can be easily processed and understood by machines or human. Currently there are several major subtasks of IE which are named entity recognition, coreference resolution, terminology extraction and, of course, relation extraction.\\ 
Recent years have seen a growing development in information extraction since 1980s. It has been widely used in many areas. For example, named entity recognition has been applied on Travel-Related Search Queries\cite{cowan2015named} and relation extraction, major research area in our project, has been applied on question answering\cite{ravichandran2002learning} and gene-disease relations\cite{jun2006extraction}. 
\section{Knowledge Base}
Knowledge base plays an pivotal role in relation extraction. Just as its name, it offers the "knowledge" for the relation extraction. The section provides an overview of knowledge base and the fundamental preliminaries for relation extraction. 
\subsection{Concept of Knowledge Base}
Knowledge base is an abstract concept. It is a universal database used to store collection of represented knowledge from specific domains in the form of logic statements.\\
Currently, there are many implementations of knowledge base. For example, one implementation of knowledge base, Freebase, represents fact using the first-logic form,"\textit{<subject> <predicate> <object>}", to define the entities and the relations between them.\\
Some implementations of knowledge bases, like Freebase, are constructed by communities manually. And others are extracted automatically from knowledge source. For instance, YAGO is generally automatically extracted from  Wikipedia, WordNet and GeoNames.\\
Knowledge base can be regarded as the input and output source of relation extraction task. Relation extraction use the knowledge representation from the knowledge base and the knowledge representation relation extraction task extracted from text or corpus can also be used to enrich the number of facts in the knowledge base. 
\subsection{Ontology}
Ontology is originally a philosophy concept which means the nature of beings, their categories and relations among the categories. In computer science, Tom Gruber defined it first as:\\
\\
\centerline{\textit{An ontology is a specification of a conceptualization}.}\cite{guber1993translational}\\
\\
This definition is difficult to understand. Let's explain it in detail. In knowledge base, an ontology is the object model with entities in specific domains and the relations hold among them. So the ontology can define the entities, the classes(types) of entities and the relations. This property of ontology gives it the ability to reason about the relationships among entities and build corresponding knowledge graph.\\
There are many representation format which can be used to describe ontology. In the early, people like to use ontology language like Cycl and KIF. Currently , the most popular data model is Resource Description Framework(RDF). Data in Freebase and YAGO database are all stored in RDF format.\\
\begin{figure}[htpb]
    \centering
    \includegraphics[width=\textwidth=6cm,height=3cm]{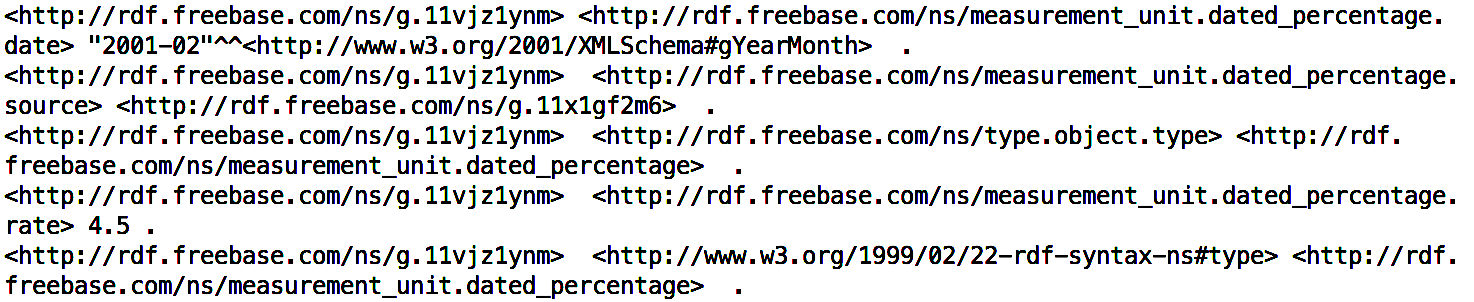}
    \caption{
        A snippet of freebase dump. 
    }
    \label{freebasedump}
\end{figure}
\subsubsection{Classes}
"Classes"(also called types, defined by different communities, for example, Freebase use "types" while YAGO use "classes"), are abstract groups which can be used to classify individuals. In Freebase, it defines "\textit{A type denotes an IS A relationship about a Topic}." For example, if we define \textit{Bill Gates} topic in \textit{Person} topic, we can say Bill Gates \textit{IS A} person. The type system is usually hierarchical. In Freebase, all the topics are the subtypes of \textit{/common/topic}. In YAGO, every entity is an instance of one or multiple classes. Every subclass, except root class, belongs to one or multiple classes as well. 
\subsubsection{Named Entity}
\label{sec:namedentity}
The term "Named Entity" comes from the Sixth Message Understanding Conference(MUC-6),1996\cite{grishman1996message}. For convenience, we call it "entity" in the next. They are the information units with entity types like names of person, location or organization and numeric values like time, date ,money\cite{nadeau2007survey}. Of course, there are more types. For instance, in our project, we extract types from Freebase which is a knowledge base hold by Google. As concluded, there are 26507 types of Freebase in August, 2015. Usually, the task to detect the entity in text is called "Named Entity Recognition(NER)". It is also a subtask of IE.
\subsubsection{Named Entity Mention}
\label{sec:mention}
The "Named Entity Mention" is the mention in the text, reference of which is a "Named Entity". Similarly, we call it "mention" in short. For example, the name "\textit{Bill Gates}" has several forms like "\textit{Bill}" or "\textit{Gates}" which can all possibly point to one entity "\textit{Bill Gates}". Linking mentions with entities needs a knowledge base like Freebase. And because of difference of the context around the mention, the link might be different. We need to know how the mentions link to entities without ambiguity. So this derived a new task named "Named Entity Disambiguation(NED)". In our project, we also used "Named Entity Disambiguation" to transform our data from YAGO format from Freebase format.
\subsubsection{Relation}
\label{sec:relation}
A "Relation" is \textit{"defined in the form of a tuple $t=(e_1,e_2,e_3,...,e_n)$ where $e_i$ are entities in a relation $r$ within document $D$"}{bach2007review}. The relation is usually binary, which means the relation only exists between two entities. But it also can be higher-order.\\
Two examples are presented below:
\begin{enumerate}[I.]
  \item Jenny and her husband Richard spoiled themselves on a mutual day off work by having
dinner at Les Bistronomes. After that they went directly back to their hotel Hyatt.
  \item At codons 12, the occurence of point mutations from G to T were observed\cite{bach2007review}.
\end{enumerate}
The expected outcomes for the first example sentence would be : marriedTo(Jenny, Richard),
dineIn(Jenny, Les Bistronomes), dineIn(Richard, Les Bistronomes), visitHotel(Jenny, Hotel Hyatt),
visitHotel(Richard, Hotel Hyatt).\\
The expected outcomes for the second example sentence would be : pointMutation(codon, 12, G, T).\\
It can be seen that in the first example, all the relations are binary while in the second example there is an 4-ary relation. Here in this project, we only focus on binary relations, which means we will only annotate corpus with the forms like marriedTo(Jenny, Richard) and dineIn(Jenny, Les Bistronomes) but not pointMutation(codon, 12, G, T).\\
\subsubsection{Resource Description Framework}
"Resource Description Framework" was originally created in 1999 on the top of Extensible Markup Language (XML) as a data model for meta data. Although there are many implementations, the model, in general, is in the \textit{<subject> <predicate> <object>} triple form. \textit{<subject>} denotes the resources and \textit{<predicate>} denotes the relationship between \textit{<subject>} and \textit{<object>}. For example, the concept "\textit{The man is a billionaire}". If we use RDF to represent it, \textit{<subject>} denotes "\textit{man}", \textit{<predicate>} denotes relationship "\textit{is-A}" and \textit{<object>} denotes "\textit{billionaire}". \\
As noticed, with RDF, it is pretty easy for RDF to represent an ontology.
\subsection{Related Work}
\subsubsection{Freebase and YAGO}
\label{freebaseyago}
Freebase and YAGO are two well-known implementations of knowledge base.\\
Freebase are extracted from Wikipedia and manually updated by communities while YAGO are automatically extracted from Wikipedia, WordNet and GeoNames. They both use RDF format to store data. Currently,Freebase has more than 23 million entities and YAGO has more than 10 million. There are cross sections in these entities because some part of them are all derived from Wikipedia. In our project, we utilized this cross section to map the entities from YAGO to Freebase.
\subsubsection{Storage Management}
For later use and the consistency of data, it should be stored in a proper database. Unlike the other kind of data, RDF data is usually stored in triple store instead of traditional relational database. Some triple store like Apache Jena is built on top of relational database. It stores data in three columns, one for "subject", one for "predicate" and one for "object". Jena use a query language named "\textit{SPARQL}", which is a RDF query language similar to SQL, to query data. Other triple store are built on their own system. For instance, Freebase is built by Metaweb using their own graph model.\\
\section{Concepts of Relation Extraction}
\label{sec:conceptofrelationextraction}
Relation extraction, as a subtask of IE, is to turn the unstructured data to structured data. However compared with other subtasks of information extraction, relation retraction focus on the relations between entities. Basically, relation extraction is a task to automatically detect and classify semantic relations between entities in text or document.\\
In relation extraction task, there are two important elements, the input of the task and approaches to get the detection and classification functions.
The input of the annotation task, as introduced above, is classes, named entity, named entity mention and relations. And the source of input is knowledge base. As mentioned above, in our project we use Freebase and YAGO to get the knowledge representation.\\ 
Currently there are four approaches to build the annotation model, supervised approaches, unsupervised approaches, semi-supervised approaches, distant supervision. They will be introduced in section \ref{sec:approaches}.
\section{Approaches}
\label{sec:approaches}
In this section, we will introduce the four approaches mentioned above. Section \ref{sec:supervisedapproaches} will introduce supervised approaches. Unsupervised approaches and semi-supervised approaches will be seen in section \ref{sec:unsupervisedapproaches}. In section \ref{sec:distantsupervisionapproaches}, we will introduce distant supervision approaches.
\subsection{Supervised Approaches}
\label{sec:supervisedapproaches}
\subsubsection{Concept}
The relation extraction can be seen as a classification task. The classifier can be trained using features extracted from the sentence with two entities and decide whether the entities are in relations or not.\\
\[
  C(F(S))=\begin{cases}
               0\\
               1\\
               2\\
               ...\\
            \end{cases}
\]
$S$ is the sentence that contains entities, $F(\cdot)$ is the function to extract a set of features from sentence $S$ involving the relation extraction tasks. $C(\cdot)$ is the classifier function to map the feature functions to different labels corresponding to relations.\\
The classifiers are usually constructed with discriminative classifiers like Softmax Regression, Support Vector Machine (SVM) or Neural Network.
\subsubsection{Discussion}
In most cases, supervised approaches can achieve high performance. However there are some serious limitations as well.
\begin{enumerate}
  \item The man labelled data is labor-consuming, resulting that the scalability of supervised approaches is not quite well. This is also the main reason we want to build a system that can handle the relation extraction task with limited data.
  \item It is difficult to extend the new entity-relation types for needs of labelled data\cite{bach2007review}.
\end{enumerate}
\subsubsection{Evaluation}
\label{evaluation}
Supervised approaches use Precision , Recall, F-measure and Accuracy to evaluate system performance.\\
\begin{equation} 
\begin{split}
\texttt{Precision} & = \frac{\texttt{Corrected extracted relations}}{\texttt{Total number of extracted relations}} \\
\texttt{Recall} & =\frac{\texttt{Corrected extracted relations}}{\texttt{Total number of relations in dataset}} \\
\texttt{F-measure} & = 2\cdot\frac{\texttt{Precision}\cdot\texttt{Recall}}{\texttt{Precision}+\texttt{Recall}} \\
\end{split}
\end{equation}
And the evaluation can also be categorized as Micro- and Macro-average of Precision, Recall and F-Measure. Macro-average gives equal weight to each relation class, while Micro-average gives equal weight to each per-relation classification decision\cite{van2013macro}.\\
For example, if there are two relations $r_1$ and $r_2$.\\
\begin{equation} 
\begin{split}
\texttt{Micro-Precision} & = \frac{\texttt{Corrected extracted relation $r_1$+$r_2$}}{\texttt{Total number of extracted relations $r_1$+$r_2$}}\\
\texttt{Macro-Precision} & = (\frac{\texttt{Corrected extracted relation $r_1$}}{\texttt{Total number of extracted relations $r_1$}}+\\
&\frac{\texttt{Corrected extracted relation $r_2$}}{\texttt{Total number of extracted relations $r_2$}})/2\\
\end{split}
\end{equation}
It can been seen Macro-precision simply divides summation of precision of $r_1$ and precision of $r_2$ by two while Micro-precision sum precision of $r_1$ and $r_2$ with different weights corresponding the number of their data points.\\
One thing should be noted that if the relation extraction task involves more than two relations, the micro-precision, recall and F-measure are the same thing. They are called the "Accuracy". The accuracy is:\\
\begin{equation}
\texttt{Accuracy} = \frac{\texttt{Corrected extracted relations}}{\texttt{Total number of all data points}}
\end{equation}
\subsection{Unsupervised Approaches and Semi-supervised Approaches}
Since this two approaches have a lot in common, we will introduce them together.
\subsubsection{Unsupervised Approaches}
\label{sec:unsupervisedapproaches}
Sometimes we have no training data and relations. In this situation, unsupervised approaches can be applied to extract new relations. There are various ways of unsupervised relation extraction. Some use clustering algorithms. Some use regular expression patterns to extract new relations. Here we present a method as a typical example described in paper \cite{banko2007open}.\\
This paper proposed a method named "TextRunner". It has three important modules,"Self-Supervised Learner", "Single-Pass Extractor", "Redundancy-Based Assessor".\\
\begin{enumerate}[I.]
  \item The Self-Supervised Learner firstly parse the sentence in corpus and then label the entity pairs $(e_i,e_j)$ as positive if the parse structure between $e_i$ and $e_j$ meets the certain constraints. This procedure will output a classifier that labels candidate extraction as "trustworthy" or not.
  \item Single-Pass Extractor extracts all possible relations and send them to classifier. In this step, only positive ones are retained.
  \item Redundancy-Based Assessor is used to rank retained relations based on a probabilistic model of redundancy in text.
\end{enumerate}
It can be seen that in this method, new relations in corpus are extracted without any hand-tagged training data and pre-defined relations. All the relations are extracted automatically.
\subsubsection{Semi-supervised Approaches}
\label{sec:semi-supervisedapproaches}
Semi-supervised approach is the combination of supervised approach and unsupervised approach.\\
In some cases, we only have few training relation tuples mentioned in \ref{sec:relation}. So tuples can be seeds and expand them to find new relations which have similar patterns with these seeds. The pattern can be generated by regular expressions or the NLP parse structure features like context words or the POS tag. And clusters can be applied here to group similar patterns. There are several well-known algorithms such as Bootstrapping\cite{hearst1992automatic} and the algorithms based on Bootstrapping like Dual Iterative Pattern Relation Extraction(DIPRE)\cite{brin1999extracting} and Snowball\cite{agichtein2000snowball}.\\
\\
As a typical algorithm, DIPRE will be an example algorithm to explain semi-supervised algorithm.\\
\begin{algorithm}
\caption{DIPRE\cite{brin1999extracting}}
\begin{algorithmic}[1]
\\Start with a small sample tuples of target relations $R'$.
\\Find all occurrences of $R'$ in document.
\\Generate patterns $P$ based on the occurrences.
\\Apply patterns to data and get more relation tuples.
\\When $R'$ is large enough, return. Else, go back to step 2.
\end{algorithmic}
\end{algorithm}
\subsubsection{Discussion}
Unsupervised and semi-supervised approaches are in fact the most popular relation system in industry currently. Because their scalability is impressive. Compared with supervised approaches, they don't need much hand-tagged training data that is hard to acquire. However they also have some shortages.\\
\begin{enumerate}
  \item For DIPRE, the pattern is difficult to be matched because it use regular expressions as patterns. So in many cases, even one punctuation can fail one match.
  \item TextRunner and Snowball have no such a problem, but they highly rely on the NER and dependency so it will be difficult for them to extend new entities and relations\cite{bach2007review}.
  \item The three of them have one common limitation which is that they could not find the relations across multiple sentences but supervised approaches could.
\end{enumerate}
\subsubsection{Evaluation}
For unsupervised and semi-supervised relation extraction, we can't not compute the precision and recall. We don't know which relation is correct and which relation is missing because relations are new. However the precision could be estimated only if we draw samples randomly from the result and check precision manually.\\
\begin{equation}
\begin{split}
\texttt{Precision} & = \frac{\texttt{Corrected extracted relations(estimated)}}{\texttt{Total number of extracted relations}} \\
\end{split}
\end{equation}
We can still not compute the recall value because the missing ones can't be estimated. The F-measure, of course, can't be computed.
\paragraph{Rand Index}
Rand index can be used to evaluate the similarity between two clusters. So if we have part of the data labelled, we can use Rand Index to measure how well our clusters fit the ground truth.\\
Given a data set of $n$ elements $S={o_1,o_2,...,o_n}$ and two partitions of $S$. $X={X_1,X_2,...,X_r}$ is a partition of $r$ subsets. $Y={Y_1,Y_2,...,Y_s}$ is a partition of $s$ subsets. Then define:\\
$a$ is the number of data points in the same set in $X$ and in the same set in $Y$\\
$b$, the number of data points in different set in $X$ and in different set in $Y$\\
Rand index $R$ is:
\begin{equation}
R=\frac{a+b}{\dbinom{n}{2}}
\end{equation}
\subsection{Distant Supervision Approach}
\label{sec:distantsupervisionapproaches}
\subsubsection{Concept}
Distant Supervision Approach is the combination of bootstrapping and the supervised approach. It, instead of using few samples as seeds, use knowledge base to acquire a large set of seeds. Then it use those seeds to create features and use supervised method to train a classifier. Here we introduce the method in \cite{mintz2009distant}.\\ 
In \cite{mintz2009distant}, the author express the intuition of his distant supervision approach is "\textit{any sentence that contains a pair of entities that participate in a known Freebase relation is likely to express that relation in some way}".\\
We can see this Distant Supervision method highly relies on Freebase. In training phase, entities are detected by Named Entity Recognition(NER). Then if one sentence contains two entities and the entities are all instances in Freebase, the feature will be extracted from sentence and added to feature vectors. Then this algorithm will train an logistic regression. In testing phase, it relies on a logistic regression classifier to predict a relation for every entity pair based on the feature vectors acquired in training phase. The features adopted in this paper are lexical features including POS tag, a sequence words between entities, syntactic features including dependency path between two entities and the named entity tag. 
\subsubsection{Discussion}
Distant Supervision Approach has both features of supervised and unsupervised approaches. It requires a lot of features and manually created knowledge similarly with supervised approaches and use very large amount of unlabelled data like unsupervised approaches. However, it doesn't need to iteratively expand patterns and it is not sensitive to genre issues in training corpus\cite{lectureA}.
\subsubsection{Evaluation}
In the paper \textit{Distant supervision for relation extraction without labelled data}, the author offered two evaluation ways, Held-out evaluation and Human evaluation\cite{mintz2009distant}. Held-out evaluation means holding out part of Freebase relation instance and compare the newly found relations with them. Human evaluation simply means letting human check if the relation exists in two entities. The latter one is much more expensive meanwhile it is more accurate.
\subsection{Hand-crafted Features of Relation Extraction}
\label{handcraftfeatures}
As we mentioned above, features are the input of learning models. There are usually two kinds of features, hand-crafted features and distributed representation features.\\
Hand-crafted features are features extracted from text with expert knowledge. They are usually in comparison with distributed representation features,which are learned with neural networks. Here, we only introduce some typical hand-crafted features and leave the introduction of distributed representation features to the next chapter.
\subsubsection{Part of Speech Tag}
Part of the speech is the tag which can mark the categories of the words. For example, in sentence \textit{He is a man}. Word "He" is a personal pronoun and its pos tag is "PRP". "ss" is a verb, 3rd person singular present. So its pos tag is "VBZ".
\subsubsection{Named Entity Tag}
As introduced in \ref{sec:namedentity}, named entity has types which can be regarded as features in relation extraction task.
\subsubsection{Context Words}
Context words are the words around the mentions.\\
For instance, in sentence \textit{"Jenny and her husband Richard spoiled themselves on a mutual day off work by having dinner at Les Bistronomes. After that they went directly back to their hotel Hyatt"}. In this sentence, \textit{Jenny} and \textit{Richard} are entities. If the size of context window is two, we can extract two words preceding and following entities as context words features. Here they are six words \textit{and}, \textit{her}, \textit{her}, \textit{husband}, \textit{spoiled},  \textit{themselves}. 
\subsubsection{Dependency Path}
\begin{figure}[htpb]
    \centering
    \includegraphics[width=\textwidth=12cm,height=4cm]{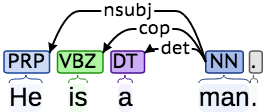}
    \caption{
        A dependency tree of a sentence generated by Stanford Core NLP online demo\protect\footnotemark .
    }
    \label{dependencypath}
\end{figure}

Dependency path is the path between words in the dependency tree of the sentence. For example, as in figure \ref{dependencypath}, the path between "He" and "man" is "He" <- "man" and the dependency tag between "He" and "man" is "nsubj". Both words and tags on the path can be features of relation extraction task. 
\footnotetext{http://nlp.stanford.edu:8080/corenlp/process}
\section{Summary}
In this chapter, we introduced the rough idea about relation extraction from a broader perspective. The concepts of information extraction were presented. We also introduced knowledge base, the input source of relation extraction. At last, four different approaches, along with the features in relation extraction task, to build models for relation extraction tasks are also described.

\chapter{Text Representation for Relation Extraction Learning}
\label{cha:textlearning}
In this chapter, text representation, as a key concept in improving the performance of our baseline relation extraction system will be introduced. Because several neural network models are pretty popular in training text representation, in section \ref{neuralnetwork}, some key points in neural network system will be explained.  We will also introduce Long-short term memory(LSTM), one kind of neural network model. Then in section \ref{sec:wordrepresentation}, the word representation including the one-hot presentation of word and distributed representation of word will be discussed. Followed is the introduction for phrase representation and sentence representation. And how phrase and sentence representation are trained by LSTM will be explained as well.
\section{Neural Network}
\label{neuralnetwork}
Artificial neural networks are a family of machine learning models used to estimate functions. In this chapter, we will introduce from a single neuron to the architecture of neural network. Then in \ref{distributedrepresentation}, how the information will be encoded in neural work will be stated. After, training approach in a neural network will be explained. At last, we will introduce some important techniques in training neural network like pre-training and fine-tuning.
\subsection{Artificial Neuron}
\label{neuron}
\afterpage{
\begin{figure}[htpb]
    \centering
    \includegraphics[width=\textwidth=2cm,height=6cm]{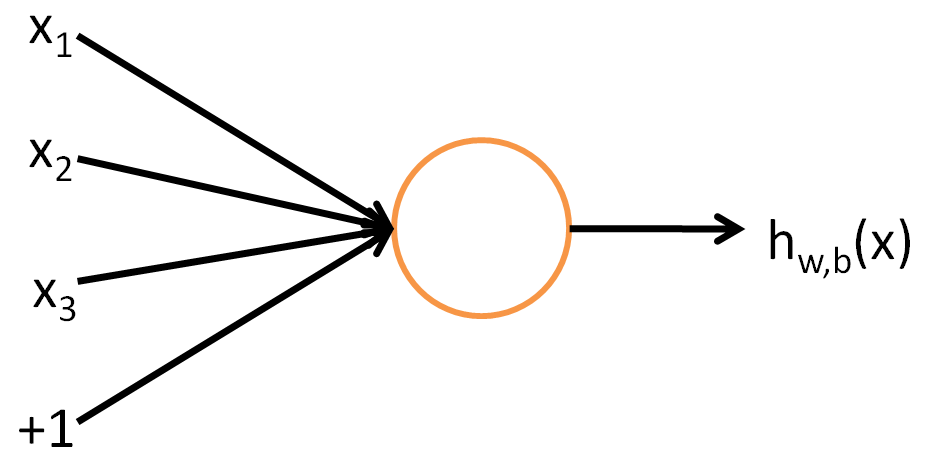}
    \caption{
        This is a single neuron\protect\footnotemark. 
    }
\end{figure}
\footnotetext{http://deeplearning.stanford.edu/wiki/images/3/3d/SingleNeuron.png}
}
Neural networks are inspired by biological neural network. Thus similarly with biological neural networks, they are composite functions composed of many single artificial neurons. An artificial neuron (also called function node) is a computation unit which can take several inputs and map them to output, $h_{W,b}(x)=f(z)=f(\sum_{i=1}^3{W_ix_i+b})$ where  $f : \Re \mapsto \Re$, $x_i$ is the input, $W_i$ is the weight on the edge, $b$ as the bias and $z=W_ix_i+b$\cite{ng2012ufldl}.\\
There are many forms of $z$. For example ,sometimes bias is added in $z$ and sometimes bias is not necessary.\\
The function $f$ in this formula is called the activation function. There are many choices of activation as well. For instance, one people often use is sigmoid function:\\
\begin{equation}
f(z)=\frac{1}{1+\exp(-z)}
\end{equation}
The characteristic of sigmoid is that it can map input to a value to $[0,1]$, which can be used to estimate the probability. So it is often applied in logistic regression as well.
The other activations include tanh functions, rectify functions or just linear functions. Depending on different situations, we choose different activation functions.\\
\subsection{Neural Network Architecture}
\label{architecture}
\afterpage{
\begin{figure}[ht]
\centering
\subfigure[Example of Feed-forward neural network\protect\footnotemark]{\label{feedforward}{\includegraphics[width=0.4\textwidth,height=0.2\textheight]{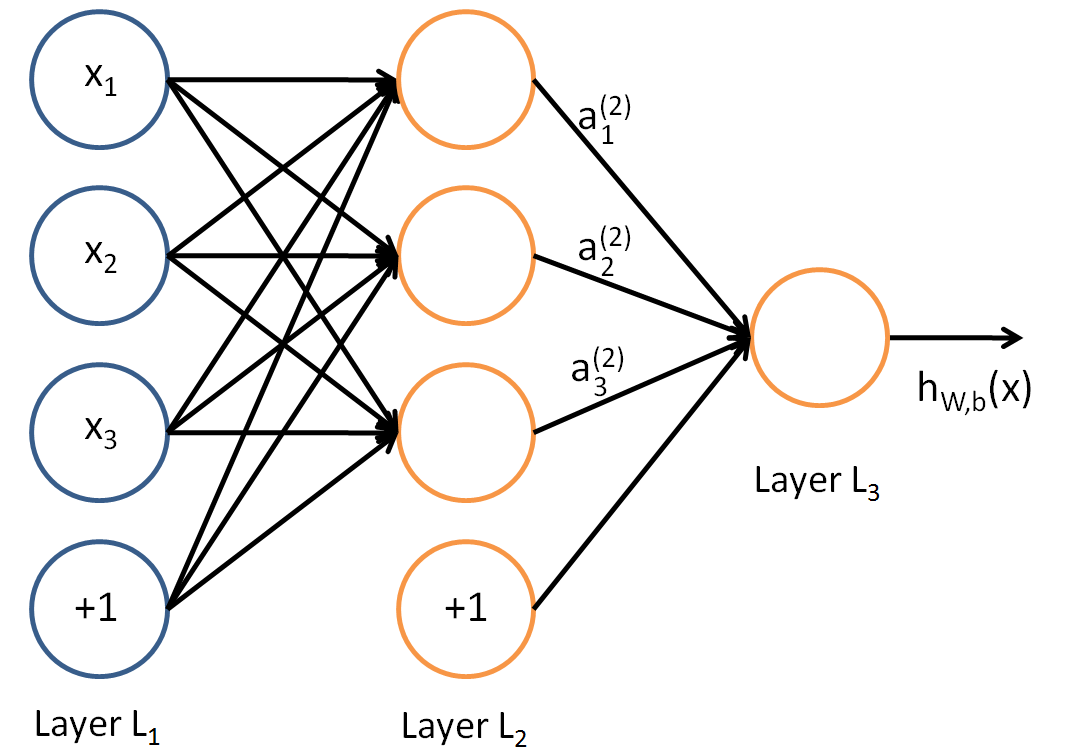}}}\hfill
\subfigure[Example of Recurrent neural network\protect\footnotemark]{\label{recurrent}{\includegraphics[width=0.4\textwidth,height=0.2\textheight]{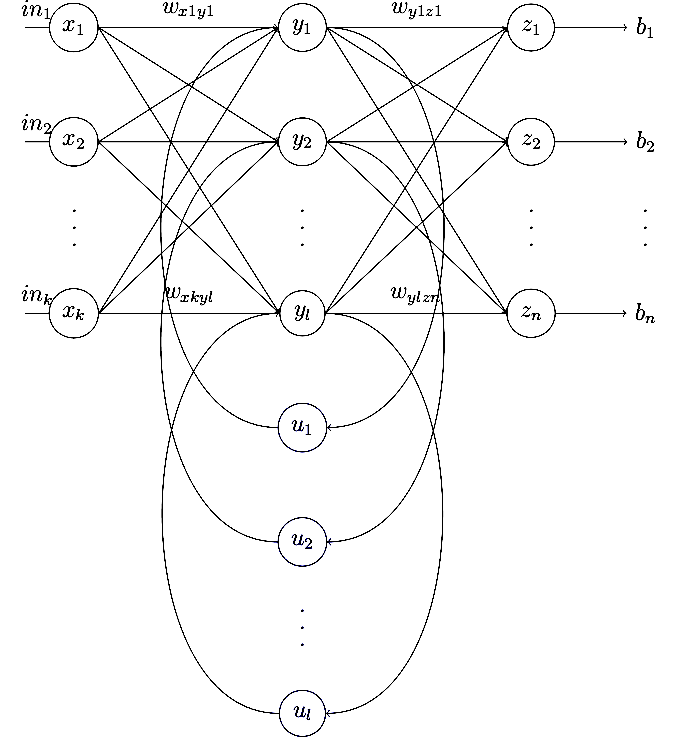}}}
\caption{Neural Network Architecture}
\label{fig:subfigures}
\end{figure}
\footnotetext{http://deeplearning.stanford.edu/wiki/images/9/99/Network331.png}
\footnotetext{https://en.wikipedia.org/wiki/Recurrent\_ neural\_ network/media/File:Elman\_ srnn.png}
}
Not only the activation functions, the structure of the neural network also varies. Usually the structure can be regarded as a tree. Different tree structures can create different neural network model. Currently, the most simple and common one is the feed-forward neural network which means the tree of the neural network has no directed cycles. If having enough hidden units, two layers neural network is capable of approximate any continuous functions to an arbitrary accuracy\cite{lectureD}.\\
As it can be seen in figure \ref{feedforward}, this is a simple feed-forward neural network model composed of 9 neuron units. There are three layers. The layer leftmost is called input layer, the layer in the mid is called hidden layer and the layer rightmost is called the output layer.\\
There are other architectures like recurrent neural network as well. On the contrary with feed-forward neural network, the tree of neural network has directed cycles. Recurrent neural network derived a lot of different kinds of neural networks. As in the figure \ref{recurrent}, it is an Elman and Jordan networks, also known as "simple recurrent networks" (SRN). And one recurrent neural network, Long short-term memory and its variants, play a pivotal role in learning text representation.
\subsubsection{Recurrent Neural Network}
\begin{figure}[htpb]
    \centering
    \includegraphics[width=\textwidth=2cm,height=6cm]{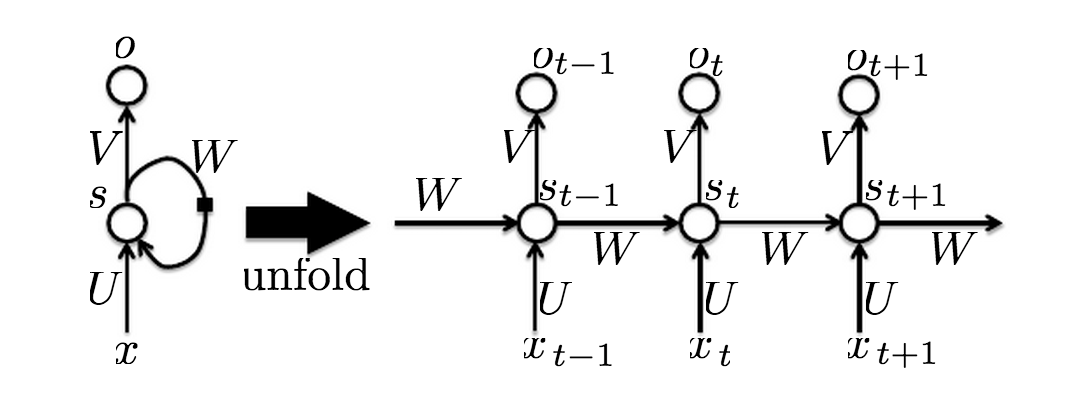}
    \caption{
        Left: Input processing part of a recurrent neural network. Right: Same seen but an unfolded graph\protect\cite{Bengio-et-al-2015-Book}. 
    }
    \label{recurrentfigure}
\end{figure}

Recurrent neural network(RNN), as I mentioned, in \ref{architecture} is the neural network with directed cycles. RNN has quite good performance on modelling sequence data because it takes the advantage of its ability of \textit{sharing parameter} across different parts of the model.\\
For example, as the figure illustrated in \ref{recurrentfigure}, $s_t= f_\theta(s_{t − 1}, x_t)$. The state $s_t$ now contains all the past sequence. The same parameters are shared over all the sequence. Since the parameter is shared over the sequence, we don't need to set different parameter for different part of the model. So the model can be generalized to different form like the length of sequence~\cite{Bengio-et-al-2015-Book}.\\
Similar with feed forward neural network, any function computable by a Turing machine can be computed by a recurrent network of a finite size\cite{Bengio-et-al-2015-Book}. 
\subsubsection{Gradient Vanishing}
As noted above, RNN is capable of making model generalize to different sequence forms. As our sentences in text are not with fixed length. Recurrent neural network has a good performance on dealing with this situation. However there is a serious problem named "\textbf{gradient vanishing}" which will influence training result of sentence representation. For example, still considering the figure \ref{recurrentfigure}, let's say:\\
\begin{equation}
\begin{split}
s_t &=\theta\phi(s_{t-1})+\theta_x x_t \\
o_t &=\theta_o\phi(h_t)\\
\frac{\partial E}{\partial \theta}
 &=\sum_{t=1}^{S}\frac{\partial E_t}{\partial\theta}
\end{split}
\end{equation}
Then in the back propagation, the chain rule is like:$\frac{\partial E}{\partial \theta}=\sum_{k=1}^{t}\frac{\partial E_t}{\partial o_t}\frac{\partial o_t}{\partial o_k}\frac{\partial o_k}{\partial \theta}$, and $\frac{\partial o_t}{\partial o_k}=\prod_{k+1}^{t}\frac{\partial o_i}{\partial o_{i-1}}=\prod_{k+1}^{t}\theta^{T}diag{\phi'(o_{i-1})}\leq\gamma_{\theta}\cdot \gamma_{\phi}$\cite{lectureC}. So if $\gamma_{\theta}\cdot \gamma_{\phi}$ is less than 1, after several transitions of state, the gradient will become close zero. Or if $\gamma_{\theta}\cdot \gamma_{\phi}$ is greater than 1, the gradient will become exponentially large. This leads to the gradient descent meaningless at some state $s_i$. After several time steps, parameters will not be updated or updated with meaningless numbers. However, to represent a sentence, it is better to have memory of the sentence. Thus, to solve the gradient vanishing, one way is to incrementally add past state to current state and the equation about the state become $s_t =1\cdot\phi(s_{t-1})+\theta_x x_t$. As we can see, the gradient vanishing problem is solved because every time the derivative equals one. However, this also brings a problem. We just simply incrementally add the last state to current state. But a good memory should not remember everything.\\
Here an alternative model named "\textbf{Long-short term memory}" could solve this problem.
\subsubsection{Long-short Term Memory}
\label{longshorttermmemory}
Long-short term memory(LSTM) is a recurrent neural network as well. LSTM has three gates,input gate, output gate and a forget gate.\\
\begin{figure}[htpb]
    \centering
    \includegraphics[width=\textwidth=3cm,height=9cm]{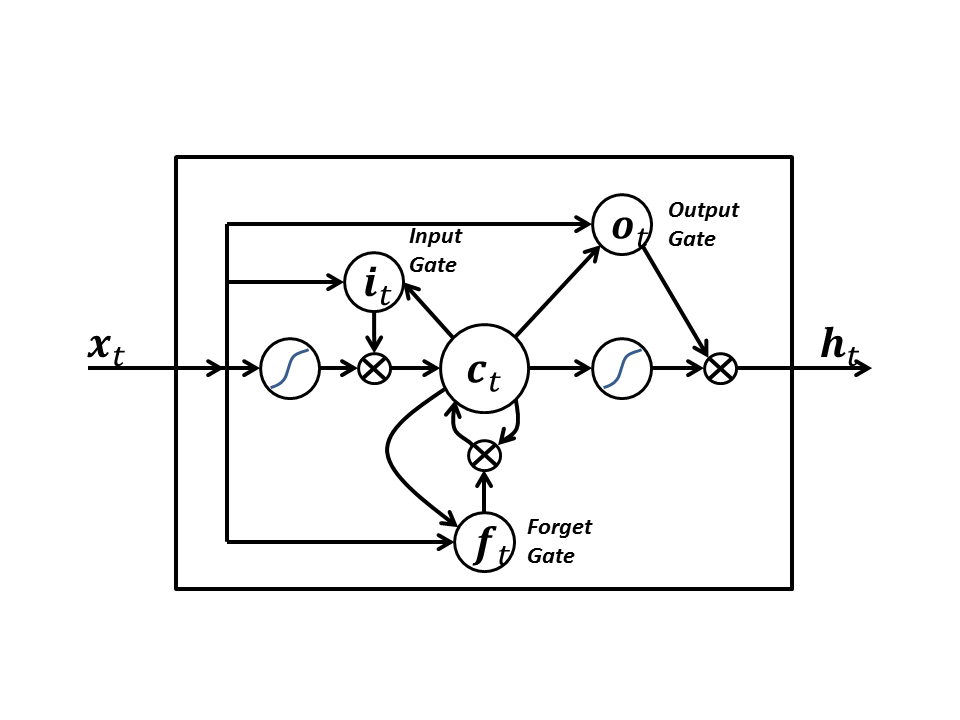}
    \caption{
        A simple LSTM cell which has input, output and forget gate\protect\cite{hochreiter1997long}. 
    }
    \label{cellLSTM}
\end{figure}
Specifically, LSTM is composed with cells and the inner structure of cell is like figure \ref{cellLSTM}. Traditional LSTM is with chain structure although there are other novel structures currently. Its transition functions are as follows\cite{hochreiter1997long}:\\
\begin{equation}
\begin{split}
i_t=\sigma(W^{(i)}x_t+U^{(i)}h_{t-1}+b^{(i)})\\
f_t=\sigma(W^{(f)}x_t+U^{(f)}h_{t-1}+b^{(f)})\\
o_t=\sigma(W^{(o)}x_t+U^{(o)}h_{t-1}+b^{(o)})\\
u_t=\tanh(W^{(u)}x_t+U^{(u)}h_{t-1}+b^{(u)})\\
c_t=i_t\odot u_t + f_t \odot c_{t-1}\\
h_t=o_t \odot \tanh(c_t)
\end{split}
\end{equation}
where $x_t$ is the input at current time step, $\sigma$ is the sigmoid function, $\odot$ is the element-wise multiplication and $W$ is the model parameters. The input gate, $i_t$, forget gate, $f_t$ and output gate, $o_t$ are in range [0,1]. Every time there is new data and past state value flowing into the cell and this three gates decide which one could pass or not. Another important thing is that output of the gate is learned by data itself. It is the data itself which decides how much supplement should be added in the flow. After this improvement, LSTM can choose to how well the last state in in the flow can influence the future state instead of simply incrementing states. Meanwhile it is able to avoid the gradient vanishing problem. This is more like a modern computer which is capable of writing, reading and erasing memory.
\subsection{Neural Network Representation for Conceptual Knowledge}
\label{distributedrepresentation}
\subsubsection{Local Representation}
In neural network, the most simple way to represent knowledge or concept is to let one neuron to represent one concept. For example, one output neuron is able to represent one cluster or one class. There are several advantages of local representation as \cite{lectureB} stated. It is easy to understand, can be coded by hand and easy to associate with other representation. However this model is inefficient because other neurons are not representing anything. And for the objects which have many properties or relations to each other, local representation has difficulty representing them. 
\subsubsection{Distributed Representation}  
To concur the limitations of local representations, the concept of distributed representation was imported. The definition of distributed representation is:\\
\theoremstyle{definition}
\begin{definition}{Distributed representation}
is a many-to-many relationship between two types of representation (such as concepts and neurons)\cite{lectureB}.
\end{definition}
It means that one neuron can represent multiple concepts and one concept can be represented by multiple neurons. Although using this kind of representation, one neuron can not have an accurate representation of each concept, the combination of several representations can be much more powerful than one local neuron using local representation. In the \ref{sec:wordrepresentation} and \ref{sentencerepresentation}, the application of distributed representation will be explained in detail.
\subsection{Neural Network Model Training}
There are two phases during training: forward propagation and backward propagation. The two phases with loss function drive the model towards our interest. In this section, we use a feed-forward neural network as an example. And we will talk about the forward propagation and loss function, then the backward propagation. In the following section, we will talk about how three components together train a complete model.
\subsubsection{Forward Propagation}
In this section, forward propagation will be stated first.\\ As introduced, the neural network is a composite of single neurons. So, in the neural network, except for the neurons in the input layer, every single neuron's input becomes the output of another neuron in the last layer. For example, in figure \ref{feedforward}, $a_1^{(2)} = f(W_{11}^{(1)}x_1 + W_{12}^{(1)} x_2 + W_{13}^{(1)} x_3 + b_1^{(1)})$ where $a_1^{(2)}$ denote the activation (output value) of unit 1 in layer 2, $f$ denotes the activation function, $W_{ij}^{(1)}$ denotes the weight which associates with the connection from unit i in layer 1 to unit j in layer 2, $x_i$ denotes the input in unit i, layer 1\cite{ng2012ufldl}. Following the same rule and computing layer by layer, we can get other output value $a_2$, $a_3$ as well. At last, after all the outputs are propagated to the output layer, we can get the final output. For example, in the figure \ref{feedforward}, since it is a pretty simple structure with only one output node, the output is  like:
$h_{W,b}(x) = a_1^{(3)} =  f(W_{11}^{(2)}a_1^{(2)} + W_{12}^{(2)} a_2^{(2)} + W_{13}^{(2)} a_3^{(2)} + b_1^{(2)})$\cite{ng2012ufldl}.\\
This procedure of mapping input representation to output representation is called forward propagation.
\subsubsection{Loss Function}
\label{lossfunction}
After forward propagation, there should be criteria to measure how well this output fits our interest. So, we need a criteria function named loss function.\\
The loss function is a concept from decision theory.
As the definition from \cite{pun2014statistical}:\\
\theoremstyle{definition}
\begin{definition}{The loss function,}
$L$ : $\Theta \times A \rightarrow R$ represents the loss when an action
$a$ is employed and $\Theta$ turns out to be the true nature of state.
\end{definition}
There are many kinds of loss functions.
For example, a commonly used
loss functions is squared error loss function. $L(\theta,a)=\frac{1}{2}(\theta-a)^2$ where $\theta$ is the ground truth and $a$ is the estimated value of $\theta$. In a neural network, $a$ is the output value computed by forward propagation. This loss function is usually used in linear regression.\\
Another commonly used loss function is logistic loss function.
$L(\theta,a)=-\sum_{i=1}^{N}\{\theta_n \ln a_i+(1-a_n)\ln(1-\theta_n)\}$ where $\theta$ is the ground truth and $a$ is the output value computed by neural network. In fact, a single neuron with sigmoid activation function and logistic loss function is the \textbf{logistic regression} model.\\
The ground truth can be manually defined by human or can be extracted from data itself. This also defines if the neural network is supervised or not. For instance, one kind of neural networks, Auto-encoder, use the input data as the ground truth. So the loss function is the error measure between the output value computed by neural network and the input value itself.\\
\subsubsection{Backward Propagation}
To minimize the loss, a common optimization method is gradient descent. Here a dynamic-programming method named \textbf{Backward Propagation} will be presented.\\
Backward propagation is a method to compute the gradient based on chain rule. There are several steps for backward propagation. First we define $a^{(l)}_i$ is the output of node $i$ in layer $l$ as we showed in \ref{neuron}. Then we define a "error term"  $\delta^{(l)}_i$ to measure how much a node is responsible for the error to the final output. Then weight on each edge is defined as $W^{(l)}_{ji}$. Then $z^{(l)}_i$ is the weighted sum of input at node $i$ in layer $l$ and $f(z^{(l)})$ is the activation function where input is $z^{(l)}$ and output is $a^{(l)}_i$. And $ L(\theta,a)$ is defined as loss function as noted in \ref{lossfunction}.Then here are the steps of backward propagation:\\
\begin{enumerate}[I.]
  \item First  calculate the partial derivative of the loss function, which is the $\delta$ of output nodes.\\
  \begin{equation}
\delta^{(n_l)}_i
= \frac{\partial}{\partial z^{(n_l)}_i} \;\;
        L(\theta,a)
  \end{equation}

  \item For hidden layers, for each node $i$ in layer $l$, we calculate the error measure $\delta$:\\
   \begin{equation}
  \delta^{(l)}_i = \left( \sum_{j=1}^{s_{l+1}} W^{(l)}_{ji} \delta^{(l+1)}_j \right) f'(z^{(l)}_i)
    \end{equation}          
  \item Calculate the the gradient using the $\delta$: \\
  \begin{equation}
  \frac{\partial L(\theta,a)}{\partial W^{(l)}_{ji}}=a^{(l)}_j \delta_i^{(l+1)}
  \end{equation}  
\end{enumerate}
From this procedure, we get gradients for every weight in the neural network.
\subsubsection{Model Parameters Learning}
\label{sec:update}
In sum, the procedure of training a neural network is to do the forward propagation first. Then use loss function to measure the error. After error is back propagated, we can get the gradients for every model parameters $W$. At last, the parameters can be updated like:
\begin{equation}
 W^{(l)}_{ji}=W^{(l)}_{ji}-\alpha a^{(l)}_j \delta_i^{(l+1)}
\end{equation}
where $\alpha$ is the learning rate. And the model is adjusted towards our learning goal.\\
Usually when learning model parameters $W$, there are some techniques which can make the performance of the model better. We will introduce them as follows.\\
\paragraph{Regularization}
When updating parameters $W$, the regularization can be added. It is a method to prevent overfitting by penalizing models with additional parameter values.\\
For example, L1 regularization is as follows:
\begin{equation}
 W^{(l)}_{ji}=W^{(l)}_{ji}-\alpha \frac{\partial L(W)}{\partial W^{(l)}_{ji}}-\alpha\lambda \text{sign} (W^{(l)}_{ji})
\end{equation}
It adds a penalty when updating model parameters.\\
The L2 regularization will modify the loss function to:\\
\begin{equation}
 \hat{L(W)}=L(W)+\frac{\lambda}{2}W'W
\end{equation}
Then when updating parameters, the gradient will become:
\begin{equation}
 W^{(l)}_{ji}=W^{(l)}_{ji}-\alpha \frac{\partial L(W) }{\partial W^{(l)}_{ji}}-\alpha\lambda  W^{(l)}_{ji}
\end{equation}
Usually, $\lambda$ is an important hyperparameter for our model. We often do \textbf{Hyperparameters Tuning} to this parameter before we officially train our model.\\
Another common regularization usually adopted by Neural Network is called \textbf{Early Stopping}. Early stopping is to test the performance of the model on the validation set after every iteration. If the performance begin to decrease instead of increasing, the training will stop.
\paragraph{AdaGrad}
For the learning rate $\alpha$, it can be a fixed number, or it can be adjusted. A common method to adjust $\alpha$ is named \textbf{AdaGrad}\cite{duchi2011adaptive}.\\
AdaGrad adjusts the learning rate component-wise based on the historical gradient. We denote the sum of squares of historical gradient as $G$, and the adjusted $\alpha$ is $\alpha_a$. A simple implementation is as follows:
\begin{equation}
\alpha_a=\frac{\alpha}{\sqrt{G}}
\end{equation}
It can be seen every time when updating parameters $W$, $\alpha$ is modified by the historical gradient.  Empirically, AdaGrad can make convergence faster and more reliable.
\subsection{Unsupervised Pre-training and Supervised Fine-tuning}
\label{sec:pretraining}
\textbf{Unsupervised Pre-training} means before doing \textbf{Supervised Fine-tuning}, we pre-trained an input representation for the following supervised task. Usually it can help improve the performance of the following supervised task because it has trained a representation as an input instead of random initialization. We can consider that random initialization almost gives zero initial information for the training the model. However as explained in \cite{erhan2010does}, there are two possibilities for the reason why pre-training helps training the neural network model. One is that pre-training can put the parameter in a proper range for further supervised training. Another one is that pre-training can initialize the model to a point in parameter space which can make the optimization process more effective. It can also be regarded as a kind of regularization because pre-training restricts parameters to particular regions\cite{erhan2010does}.\\ 
However the training strategy for pre-training is usually greedy but not based on the global interest. Thus the representation is only sub-optimal\cite{Bengio-et-al-2015-Book}. So this is the reason that, after pre-training, there must be a phase named \textbf{Supervised Fine-tuning}. Fine-tuning is to train a model to fit the global interest.
\\After pre-training and fine-tuning, the model's performance is empirically better than the methods using only supervised training.
\section{Word Representation}
\label{sec:wordrepresentation}
We introduced local representation and distributed representation in \ref{distributedrepresentation}. Here their application on words will be introduced respectively.
\subsection{One-hot Word Representation}
Conventionally, supervised lexicalized NLP approaches take a word and convert it to a symbolic ID, which is then transformed into a feature vector using a one-hot representation\cite{turian2010word}.\\
In relation extraction one of the most common features is the context words of entities. For example, in the sentence "\textit{Jenny and her husband Richard spoiled themselves on a mutual day off work by having dinner at Les Bistronomes. After that they went directly back to their hotel Hyatt.}", we can extract two proceeding words and following words around the entities. Here for relation dineIn(Richard, Les Bistronomes) the context words are "her","husband","spoiled","themselves" around "Richard" and "dinner", "at" around "Les Bistronomes"(because there are no words after "Les Bistronomes", so there are only two words around "Les Bistronomes").\\
Then we can transform it to one-hot word representation. One-hot representation is a kind of local representation. It encodes concept or knowledge using 1 or 0. The one-hot representation is like:\\
\begin{align*}
\textbf{her} && (1,0,0,0,0,0)\\
\textbf{husband} && (0,1,0,0,0,0)\\
\textbf{spoiled} && (0,0,1,0,0,0)\\
\textbf{themselves} && (0,0,0,1,0,0)\\
\textbf{dinner} && (0,0,0,0,1,0)\\
\textbf{at} && (0,0,0,0,0,1)\\
\end{align*}
It can be seen that all the words are converted to vector-like features with only 1 and 0. \\
However, this method to represent word has some serious limitations:
\begin{enumerate}[I.]
  \item The one-hot representation is usually high-dimensional which is computationally expensive for some specific algorithms
  \item This kind of representations do not generalize well. For example, it is not capable of representing the relations between words in vector space. Every word in vector space is equal to each other.
\end{enumerate}
Because of these limitations, under certain situations, distributed representation of words can be used to improve the performance of relation extraction task.
\subsection{Distributed Representation of Word}
Distributed representation of word is a kind of representations of words which can help improve the performance of natural language processing task by grouping similar words in the form of word vector.\\
\afterpage{
\begin{figure}[htpb]
    \centering
    \includegraphics[width=\textwidth=3cm,height=9cm]{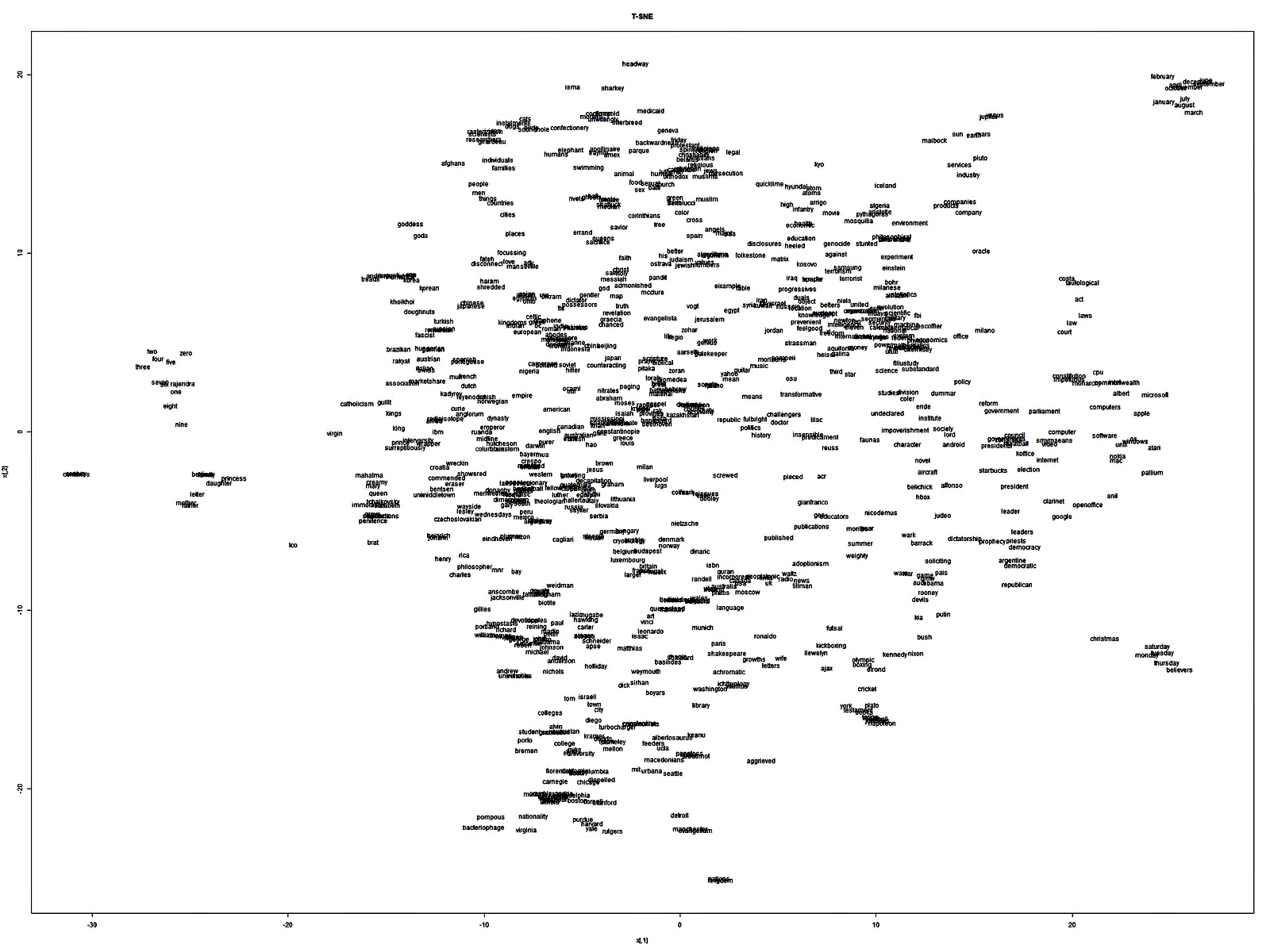}
    \caption{
        This is an example plot of word representation\protect\footnotemark. 
    }
\end{figure}
\footnotetext{http://4.bp.blogspot.com/-jGdnqFCJsP8/U6XQ3tLJrQI/AAAAAAAAIE0/w8lfdAhcRRk/s1600/plot.5.jpg}
}
For example, the distance between different words follow the specific rules in geometry space:\\
\begin{align*}
W("woman") - W("man") &\simeq W("aunt") - W("uncle")\\
W("woman") - W("man") &\simeq W("queen")  -  W("king")
\end{align*}
We can see the distance between word \textit{woman} and word \textit{man} is very close to the distance between \textit{aunt} and \textit{uncle}. This appearance displays that the distributed representation of words learned with particular training criteria owns the ability to group words. The technique mapping words to distributed representation in the form of the vector is also called \textbf{Word Embedding}.
\subsubsection{Skip-gram Model}
\label{skipgram}
Now a common model to learn this kind of representation is called Skip-gram model\cite{mikolov2013distributed}. To get a representation of a word, we must maximize the likelihood below:
\begin{equation}
\frac{1}{T}\sum_{t=1}^{T}\sum_{-c\leq j \leq c,j\neq 0}\log p(w_{t+j}|w_t)
\label{skip-gram}
\end{equation}
where $c$ represents the training context size, which may vary based on the central word $w_t$, and $w_t$ denotes the words in the training sequence.\\
The $p(w_{t+j}|w_t)$ is defined using softmax:\\
\begin{equation}
p(c|w) = \frac{\exp(v_c \cdot v_w)}{\sum_{c'\in C}\exp(v_c' \cdot v_w)}
\end{equation}
where $c$ is the context word and $w$ is the target word. And $C$ is the vocabulary set with all possible context words. As we can see that this formula can't be implemented in practical situation because the $C$ is too large with $10^5$ - $10^7$ terms\cite{mikolov2013distributed}. So another approach must be used.
\paragraph{Negative Sampling}
To maximize the log likelihood \ref{skip-gram}, a common way is negative sampling. Method is as the following steps:\\
\begin{enumerate}[I.]
  \item First label the origin sequence around the target word as true.
  \item Random sample noise word $k$ times from the word repository to replace with one context word of target words.
  \item Label the sampled sequence as false.
  \item Then replace $p(w_{t+j}|w_t)$ with
\begin{equation}
	\log\ \sigma(v_c\cdot v_w) + \sum_{i=1}^{k}E_{w_i \sim P_n(w)}[\log\ \sigma(-v_{c'_i} \cdot v_w)]
\end{equation}
where $v_w$ is the target word, $v_c$ is the vector of context word being replaced, $k$ is the times of negative sampling for $v_c$. And $v_{c'_i}$ is the noise word. $v_c\cdot v_w$ is what we call the score function, simply element-wise multiplication of the two vectors.\\
\end{enumerate}
Although there are many other models like Continuous Bag-of-Words(COW) Model\cite{mikolov2013efficient}, in our project we use the word2vec which is the vector collection trained by Skip-gram model.
\subsubsection{Summary}
In skip-gram model, the representation is grouped with its semantic meaning because its training object is to maximize the likelihood of its context words given the target words. This model is trained with documents in Wikipedia. However, this training object is not fixed. For example, the COW model's training goal is to maximize the likelihood of the target word given its context word.\\
Different training goal give different performance in different tasks. In fact, different training goal defines the distance between words in its own criteria so the vectors are grouped by corresponding rule in vector space. And in our project, we will define our own distance to fit our requirement.
\section{Representation of Phrase and Sentence}
In this section, we will focus on how to acquire the representation of phrase and sentence. 
\subsection{One-hot Representation of Phrase and Sentence}
Hand-crafted features could also be encoded with one-hot representation. For example, the simple unigram feature of a sentence could be encoded with a vector with the size of vocabulary. All the words in the sentence will be encoded with 1s otherwise 0s. The approach for sentence modelling usually doesn't generalize well when training data size is small.
\subsection{Distributed Representation of Phrase and Sentence}
There are several typical methods to acquire the distributed representation of phrase and sentence. We will focus on the method of LSTM. Other typical methods will also be introduced.
\subsubsection{Distributed Representation of Phrase and Sentence Learning with LSTM}
\label{sentencerepresentation}
We can use LSTM to train a sentence or phrase representation. For different tasks, we can choose different training criteria. And based on the different training criteria, the representation LSTM learned will be different.\\
For instance, in \cite{sutskever2014sequence}, their task is to do machine translation. Thus their strategy is to map the input sequence to a fixed-sized vector using LSTM. Then the learning goal of their the model is to maximize the $p(y_1,..., y_{T'} |x_1, . . . , x_T )$ where $x$ is the input sequence and $y$ is the output sequence. The length of input and output might be different. Here:
\begin{equation}
p(y_1, . . . , y_{T'} |x_1, . . . , x_T )=\prod_{t=1}^{T'}p(y_t|v,y_1,...,y_{t-1})
\end{equation}
where $v$ is the fixed dimensional representation of the input sequence $(x_1,...,x_{t-1})$ give n by last hidden state of LSTM and each $p(y_t|v,y_1,...,y_{t-1})$ is regarded as a softmax over all the vocabularies\cite{sutskever2014sequence}.\\
\subsubsection{Typical Approaches}
\label{typicalmethods}
Modelling sentences and phrases with LSTM is not the only method. In fact, we could use any neural network models to model sentences and phrases. But the quality of different neural sentence models will be different. Empirically, LSTM is a relatively better model than other ones. Except modelling with neural networks, there are other typical methods proposed. We will introduce them in the following.
\paragraph{Average Vectors} Both phrase and sentence can be acquired by simply weighted average the distributed representation of words in sentence\cite{le2014distributed}. But this method doesn't consider the order between words.
\paragraph{Concatenate Vectors} Sentence representation can also be acquired by concatenate the distributed representation of words in an order given by the parse tree of a sentence. But this method highly relies on parse. So it is only for sentences.
\paragraph{Skip-gram Model for Learning Phrase} Phrase can also be trained using skip-gram model as well just like the distributed representation of words. However there is a little difference. There is a score function used to form a phrase which is like:
\begin{equation}
score(w_i,w_j)=\frac{count(w_iw_j)-\delta}{count(w_i)\times count(w_j)}
\end{equation}
where the $\theta$ is the discounting coefficient used to prevent too many phrases consisting of infrequent words. Here it can be seen that, compared with the score function of word representation, this score function is formed based on the count of unigram and bigram. 
\section{Summary}
Text representation is a broad field which includes a lot of methods. In this chapter, we just introduced a few examples for our convenience to present our own ideas to represent text. In the next chapter, we will focus on how to construct our own text representation model and use it to improve the performance of our baseline system.

\chapter{Models for Extracting Relations}
In this chapter, we will introduce the baseline models in \ref{baseline}. Then our new models for extracting relations will be proposed. It should be mentioned that our system along with baseline only focus on binary relations. Thus in our work, the higher-order relation won't be considered.\\
\section{Baseline System with Hand-crafted Features} 
\label{baseline}
In this section, the details about the baseline system with hand-crafted features will be proposed. In \ref{overview}, the broad concepts about our baseline will be introduced. Then in \ref{feature}, the features we selected for the baseline system will be explained.
\subsection{Overview}
\label{overview}
\begin{figure}[htpb]
    \centering
    \includegraphics[width=50mm,scale=0.5]{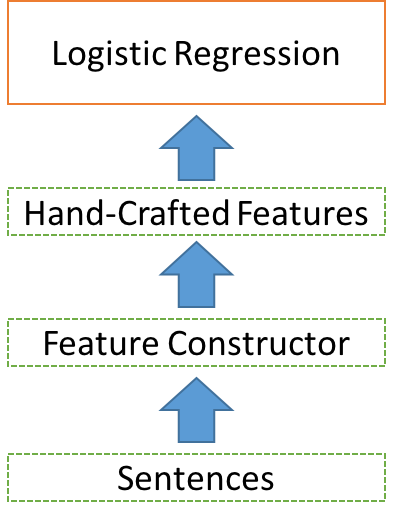}
    \caption{
        Architecture of Baseline System with Hand-crafted Features.
    }
    \label{baselinearchtecture}
\end{figure}

Our baseline is derived from two papers \cite{chan2010exploiting} and \cite{chan2011exploiting}.
For the baseline system, our model adopt logistic regression model with the input of traditional hand-crafted features. As we can see, in figure \ref{baselinearchtecture}, the hand-crafted features will be extracted from sentences and encoded with one-hot representation. Then it will be the input of logistic regression. The optimization algorithm in baseline is Adagrad, which has been introduced in \ref{sec:update}.\\
At the end of this project, the performance of this baseline system will be the main comparison for the performance of new system with unsupervised pre-training representation learning and supervised fine-tuning phase.
\subsection{Feature Selection}
\label{feature}
For the baseline, we use five types of features selected from \cite{chan2010exploiting} and \cite{chan2011exploiting}.
\begin{enumerate}[I.]
  \item \textbf{Base chunk feature:} Base chunk feature is a boolean feature to see whether there is a phrase chunk between two mentions.
  \item \textbf{Collocations feature:} Collocations feature is a set of features denoted by $C_{i,j}$. For example, $C_{-1,+1}$ is a sequence of three tokens, the token on the immediate left of the head word of an entity, the token of this head word and the token on the immediate right of head word. Here we use the $C_{-1,-1}$,$C_{+1,+1}$,$C_{-2,-1}$,$C_{-1,+1}$ and $C_{+1,+2}$.
  \item \textbf{Syntactic parse feature:} Syntactic parse feature is tokens and part-of-speech tags on dependency path between two entities.
  \item \textbf{Lexical feature:} Lexical feature contains a set of features. It includes head word of $m_1$ and head word of $m_2$ and head words of $m_1$ and $m_2$ combining together where $m_1$ is the first mention in sentences and $m_2$ is the second one. Other lexical features are tokens in mentions, unigram between mentions and bigram between mentions.
  \item \textbf{Dependency feature:} Dependency feature is also set of features.It includes head of $m_1$ and its dependency parent, head of $m_2$ and its dependency parent, the words on dependency path between two mentions and dependency labels on the path.
\end{enumerate}
\section{Novel Representation Learning Model}
In this section, we will introduce our representation learning models. Firstly we will explain the preliminary background including the motivation behind our models and the novel structure of LSTM we used. Then we will propose the architecture of our system. At last, we will introduce our approach to learn model parameters.
\subsection{Motivation}
\begin{figure}[htpb]
    \centering
    \includegraphics[width=\textwidth=3cm,height=8cm]{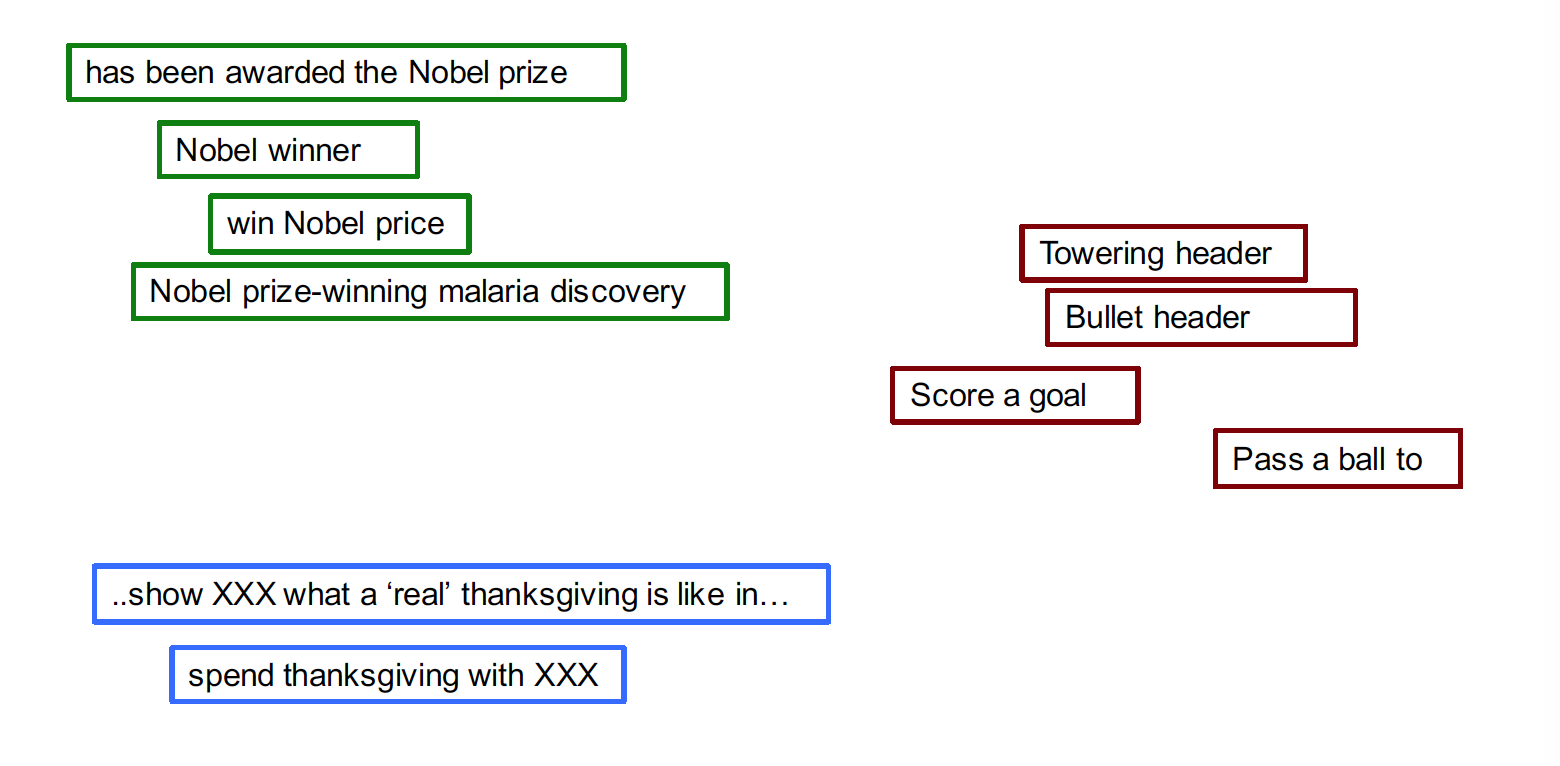}
    \caption{
        Expression grouped by its relation.
    }
    \label{expression}
\end{figure}

In baseline system, we use one-hot representations and logistic regression. Because of paraphrase problem -- one relation may be expressed in many ways, the number of dimensions will be pretty high. And also since supervised approaches need manually labelled data, we can only get limited training data in most cases. However, logistic regression has a lousy performance with limited training data but high dimension features. To solve those problems, we could use distributed representations.\\
It is intuitive that the semantic expressions around two entities are highly correlated with the relation between two entities. If we can map expressions into latent distributed representations and the representations labelled with same relations can be grouped together such as in the figure \ref{expression}, it will be easy for logistic regression to find optimal decision boundaries for classifications of relations.\\
\subsection{Shortest Dependency Path LSTM}
In out project, we adopt a novel LSTM model, \textbf{Tree LSTM}\cite{tai2015improved}, to map the expression to latent distributed representations. And we choose \textbf{Shortest Dependency Path} as the semantic expression.
\subsubsection{Tree LSTM}
\begin{figure}[htpb]
    \centering
    \includegraphics[width=\textwidth=3cm,height=8cm]{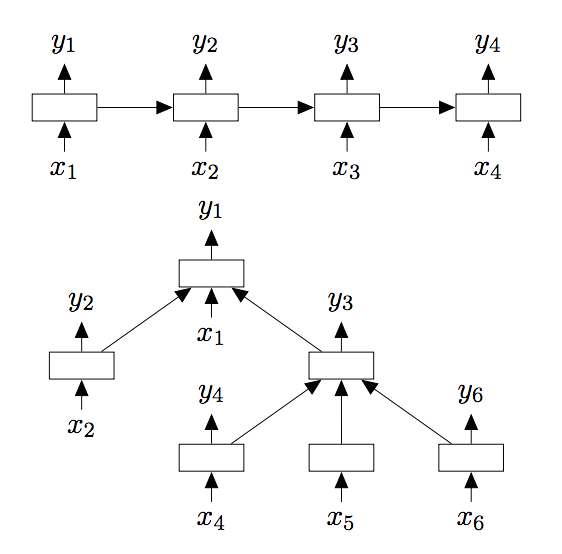}
    \caption{
        \protect\textbf{Top:} a traditional chain LSTM structure. \protect\textbf{Bottom:} a tree LSTM structure\protect\cite{tai2015improved}. 
    }
    \label{lstmarchitecutre}
\end{figure}

The transition equations of this binary LSTM are as follows:
\begin{equation}
\begin{split}
i_t&=\sigma(W^{(i)}x_t+U^{(i)}h_{t-1}+b^{(i)})\\
f_t&=\sigma(W^{(f)}x_t+U^{(f)}h_{t-1}+b^{(f)})\\
o_t&=\sigma(W^{(o)}x_t+U^{(o)}h_{t-1}+b^{(o)})\\
u_t&=\tanh(W^{(u)}x_t+U^{(u)}h_{t-1}+b^{(u)})\\
c_j&=i_j\odot u_j + \sum_{l=1}^{N}f_{jl} \odot c_{jl}\\
h_t&=o_t \odot \tanh(c_t)
\end{split}
\end{equation}
where $x_t$ is the input at current time step, $\sigma$ is the sigmoid function, $\odot$ is the element-wise multiplication and $W$ is the model parameters. The transition functions are almost as the traditional LSTM. The only difference is the transition function, $c_j=i_j\odot u_j + \sum_{l=1}^{N}f_{jl} \odot c_{jl}$, where $l$ is the sub-level of $j$ and $c_j$, becomes summation of its children cells' output instead of only the last cell. The comparison between chain structure and tree structure is as the figure in \ref{lstmarchitecutre}.\\\\
\paragraph{Discussion}
The reason that we use LSTM is that, as noted in \ref{longshorttermmemory}, LSTM has a good performance on sequence data and it won't have the gradient vanishing problem like normal recurrent neural networks. And also this binary tree structure is good at catching the relation patterns in the expressions.
\begin{figure}[htpb]
    \centering
    \includegraphics[width=\textwidth=8cm,height=4cm]{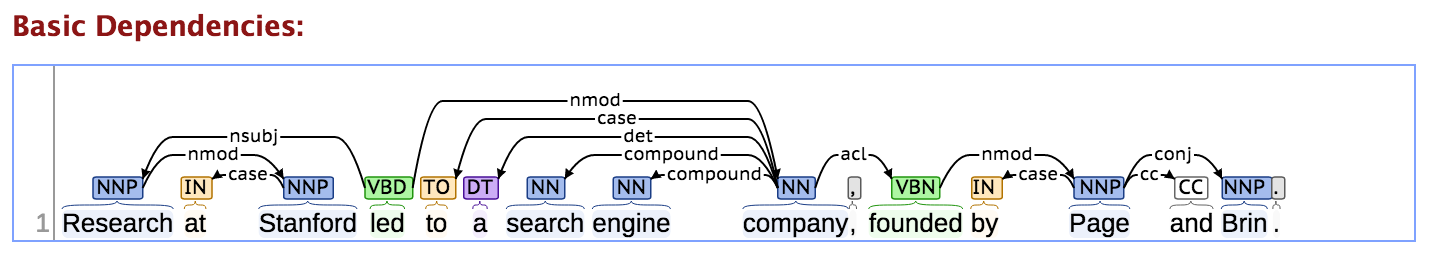}
    \caption{
        This is a dependency tree parsed by Stanford Core NlP online demo\protect\footnotemark. 
    }
    \label{stanfordnlp}
\end{figure}
\footnotetext{http://nlp.stanford.edu:8080/corenlp/process}
\subsubsection{Shortest Dependency Path} 
\begin{figure}[htpb]
    \centering
    \includegraphics[width=\textwidth=3cm,height=8cm]{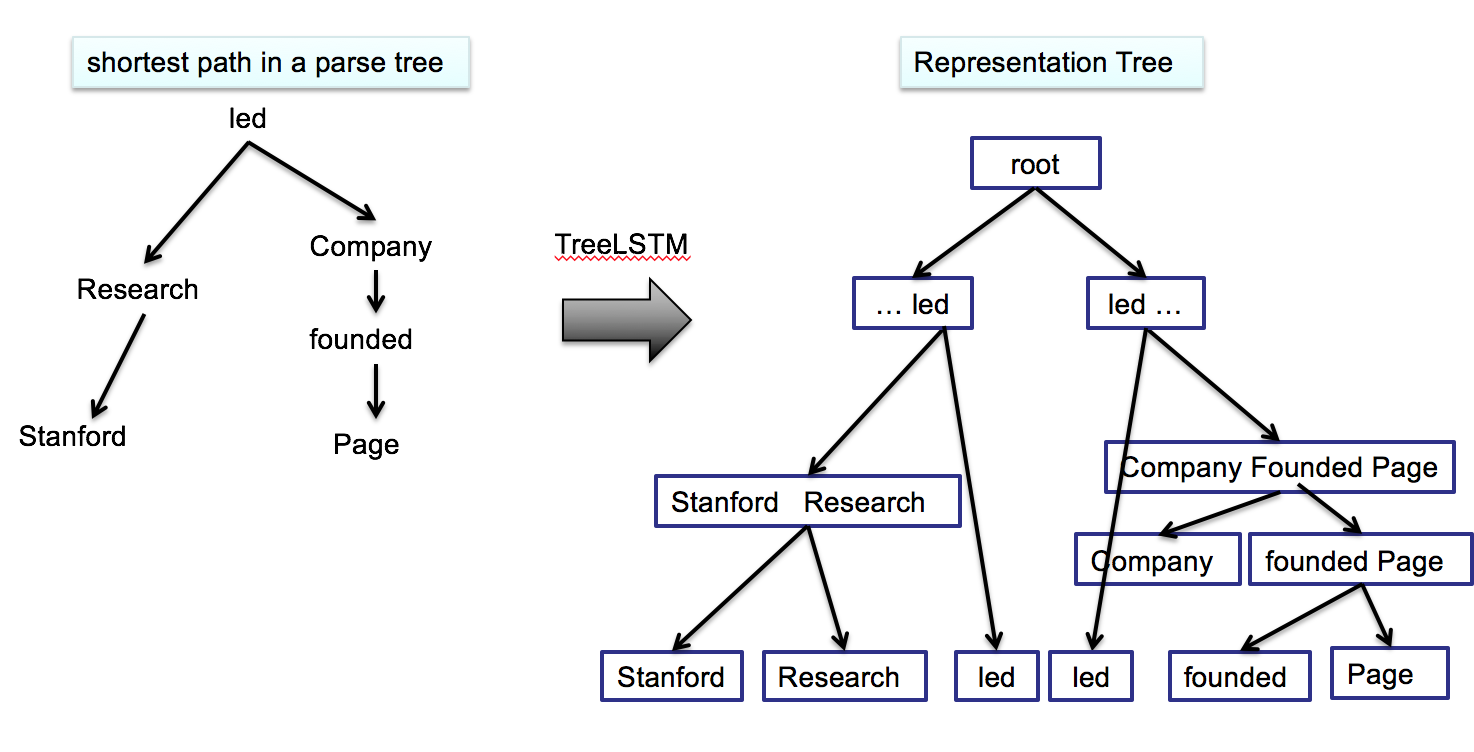}
    \caption{
        Shortest dependency path to LSTM tree.
    }
    \label{pathtotree}
\end{figure}

The expression we decide to use is the shortest dependency path between two mentions $m_i$ and $m_j$ in the sentence linking with entity $e_i$ and entity $e_j$. Shortest dependency path is the shortest path between two words in the dependency tree. For example, as in figure \ref{stanfordnlp}, the path between words "Stanford" and "Page" is "Stanford" <- "Research" <- "led"(ROOT) -> "Company" -> "founded" -> "Page" where "led" is the root of the dependency tree. \\
It should be noticed that the mentions may contain multiple words. So here we use path between head words $hw$ of mentions $m$ as the path between mentions. To get the head word, we use one heuristic method from \cite{chan2011exploiting}: if mention $m$ contains a preposition and a noun preceding the preposition, the noun will be the head word $hw$. If there is no preposition in $m$, the last noun in $m$ will be the $hw$.\\
Here we ignore the direction in the dependency path, but the root should be marked because it will be of use in building the LSTM tree.\\ 
Here, following the shortest dependency path, we construct the tree LSTM as the following steps. The input for every node in the input layer is the entity vector and word vector. Then we merge the node from mentions to root. Because there are only two mentions, we will get two branches. At last, we merge those two branches and get a complete tree. For example, as in figure \ref{pathtotree}, "Stanford" merges with "Research" to get node with "Stanford Research". Then node with "Stanford Research" merge with root "receive" and we get the left branch. For the right branch, "Page" merges with "founded", then "founded Page" merges with "Company", at last merges with the root "receive". Finally left and right branches can be merged to one completed tree. 
\paragraph{Discussion} The reason that we use shortest dependency path is, empirically, the words on dependency path between two mentions contain key information about relations between entities. And the shortest path guarantees us to only get the most important information so that we don't have to take too much computation when training the model.
\subsection{Architecture of Novel System}
\begin{figure}[htpb]
    \centering
    \includegraphics[width=\textwidth=3cm,height=10cm]{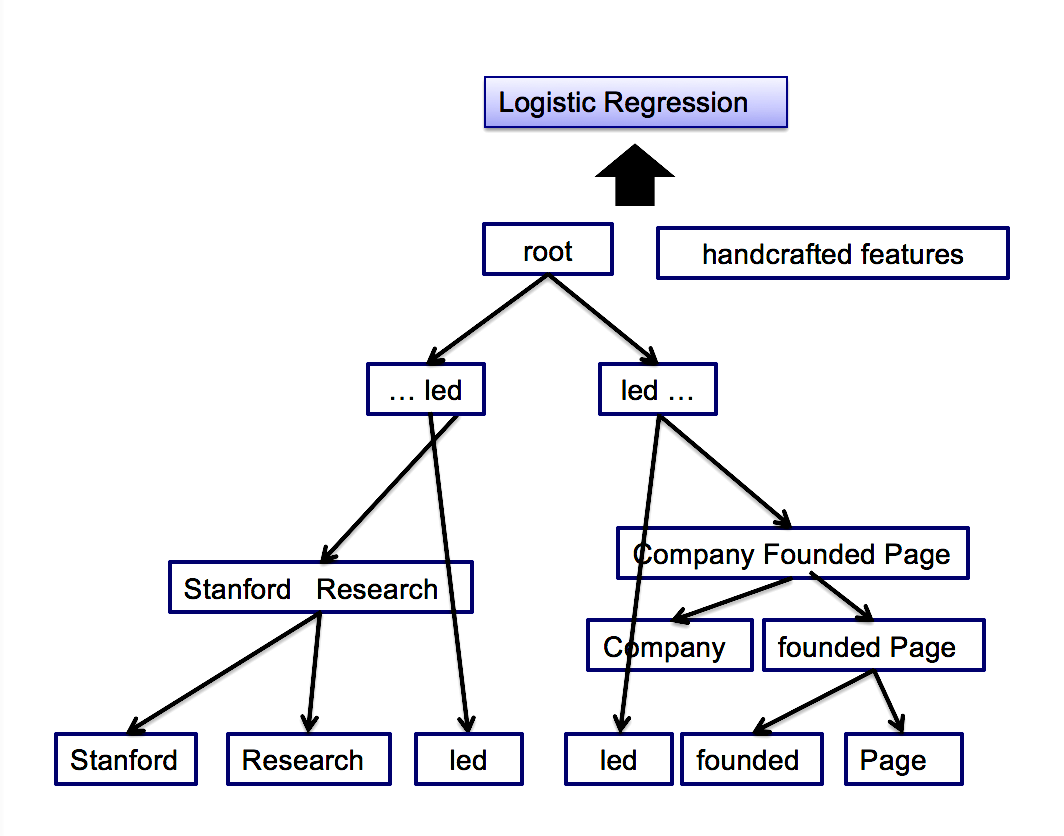}
    \caption{
        The process of fine-tuning.
    }
    \label{fintuning}
\end{figure}

The architecture of our novel system is as figure \ref{fintuning}. After LSTM mapping the expressions to latent representation, we input both hand-crafted feature and expression representations to logistic regression. After training, we will get our model parameter.\\
However, there are two choices for the parameter initialization of LSTM.\\
First is the random initialization. In our project, we just consider it as another baseline system used to evaluate whether pre-training is helpful.\\
In another option, the model parameters of LSTM are learned during pre-training. In this method, there are two phases, \textbf{pre-training} and \textbf{fine-tuning}. Pre-training, as introduced in \ref{sec:pretraining}, is able to initialize the model to a point in parameter space which can make the optimization process more effective\cite{erhan2010does}. For our task, after pre-training, LSTM is able to map expressions to latent representations. Under optimal situations, the representations can be pre-grouped based on their types of relations. \\
Because pre-training is a greedy strategy, fine-tuning can help us train a model which fits our global interest. For our project, fine-tuning will combine the pre-trained latent representations and hand-crafted features together to improve the performance of baseline.\\
\subsection{Model Parameters Learning}
In this section, we will introduce how to learn the parameters of our models. Firstly methods to initialize the input of our models will be proposed. Then we will introduce the pre-training phase including the loss functions and how we update parameters. At last, the fine-tuning phase will be introduced.
\subsubsection{Word and Entity Representations Initialization}
Before pre-training and fine-tuning, we need to build an entity vector repository $E_{e}$ and a word vector repository $E_{w}$ as the input for models.\\
For normal words, we use the word2vec repository, which is a collection of distributed representation of words trained by Skip-gram model, to map words in corpus to word vectors.\\
For entities, since entities usually have many surface patterns, we need to do some pre-processing. As mentioned in the \ref{typicalmethods}, some people average word vectors for words in the phrase or sentence to create phrase or sentence representations. So here we will use a similar approach to create our representations for entities.\\
Our strategy is to average the word vectors in all mentions of one entity to create the entity vector. However, multiple mentions can be mapped to one entity. So one way is to go through all the sentences in the dataset and find the all mentions linking to the entity. Then word vectors in mentions for each entity will be averaged. 
\subsubsection{Pre-training}
With the input, we could begin to learn the parameters of LSTM. There are two loss functions in our pre-training phase.\\
One loss function is at the output node of LSTM, used for estimating the conditional probability of entities given the shortest dependency path representation. Because the loss can be considered as using the shortest dependency path representation to predict entities around, we refer it as \textbf{Entity Prediction Loss Function} $L_o$.\\
Another loss function is at inner nodes of LSTM, used for estimating the conditional probability of local context words given the phrase representation generated on each inner node. Because the loss can be considered as using the phrase representation to predict local context words, we refer it as \textbf{Word Prediction Loss Function} $L_i$. This loss function is optional. We will evaluate the performance of this loss function in a series of experiments.\\
As noted, our interest of pre-training is to train a model which is able to map expressions to the latent representations that can be pre-grouped based on their relations. To explain the reason that the two loss functions fit our interest, the loss functions, as well as the motivations behind them, will be explained.
\paragraph{Entity Prediction Loss Function}
Entity Prediction loss function $L_o$ is:
\begin{equation}
\log \sigma (s(e_i,e_j,f(x_{ij}))+ \sum_{i=1}^{k}E_{e_i \sim P(e_i)}\log \sigma(-s(e_i,e_j,f(x_{ij})))+\sum_{i=1}^{k}E_{e_j \sim P(e_j)}\log \sigma(-s(e_i,e_j,f(x_{ij})))
\label{equation1}
\end{equation}
where $e_i$ and $e_i$ are two entities in the sentence and $f(x_{ij})$ is the shortest dependency path representation between $e_i$ and $e_j$.\\
This loss function is at output node. It is for measuring the correlation between expression representation and the entities around.\\
The idea of this loss function is borrowed from the Skip-gram model \cite{mikolov2013efficient}. Looking back the Skip-gram model introduced in \ref{skipgram}, we find that the key component of Skip-gram is to maximize scores computed by taking the dot product of pairs of latent vectors for the context words. Take a close look at the dot product of two vectors in Skip-gram score function,
\begin{equation}
s=v_i'\cdot v_j\\
\end{equation} 
The partial derivative of $s$ is w.r.t $v_i$ and $v_j$ are $v_j$ and $v_i$ respectively. If we update parameters with stochastic gradient descent, 
\begin{equation}
v_i^{t+1}=v_i^{t} + v_j^{t} 
\end{equation}
\begin{equation}
v_j^{t+1}=v_j^{t} + v_i^{t}
\end{equation}
The element $v_i^{t}$ and $v_j^{t}$ will enhance each other if they share the same sign or they will go to zero together. $v_i$ and $v_j$ are each other's context word. Thus the Skip-gram model is able to encode the information of context words into the target words. The result is that if two words have the similar context words around, their word vectors trained by Skip-gram model will be close to each other in the feature space.\\
The method we use to achieve our goal is similar to Skip-gram model's. We have a hypothesis:\\
\begin{hyp}
If two expressions have the same entities around them, there is a high possibility that they express the same relation. 
\end{hyp}
In order to group the expressions, we can maximize the likelihood:\\
\begin{equation}
\frac{1}{I}\cdot \frac{1}{J}\sum_{i=1}^{I}\sum_{i=1}^{J}(\log p(e_i|x_{ij},,e_j)+\log p(e_j|x_{ij},,e_i))
\label{equation2}
\end{equation}
where $e_i$ and $e_j$ are two entities and $x_{ij}$ are the shortest dependency between $e_i$ and $e_j$.\\
Then we can design a score function, similar with Skip-gram's, to capture the co-occurrence patterns of entities and their relational expressions. If we start with entity and word vectors, then we can map the expression $x_{ij}$ around two entities $e_i$ and $e_j$ with the map function $f(x_{ij})$ trained by LSTM. Then we can get our own score function:
\begin{equation}
s=e_i' \cdot diag(f(x_{ij})) \cdot e_j
\end{equation}
This refers to the the dot product multiplication of this three vectors and we can get a scalar number as the score. The likelihood defined by score function and softmax function is:\\
\begin{equation}
\frac{1}{I}\cdot \frac{1}{J}\sum_{i=1}^{I}\sum_{i=1}^{J}(\log \frac{s(e_i,diag(f(x_{ij})),e_j)}{\sum_{e_i \in E_e}s(e_i,diag(f(x_{ij})),e_j)}+\log \frac{s(e_i,diag(f(x_{ij})),e_j)}{\sum_{e_j \in E}s(e_i,diag(f(x_{ij})),e_j)}
\end{equation}
where $E_e$ is the entity repository.\\
As we can see, maximization of this softmax function has the same problem as when we are maximizing the softmax function of Skip-gram model. This formulation is impractical because the number of entities is too large. So the negative sampling method is adopted. The loss function becomes:\\
\begin{equation}
\log p(e_i|x_{ij},e_j)=\log \sigma (s(e_i,e_j,f(x_{ij}))+ \sum_{i=1}^{k}E_{e_i \sim P(e_i)}\log \sigma(-s(e_i,e_j,f(x_{ij})))
\end{equation}
where $E_{e_i \sim P(e_i)}$ means the sample expectation.\\
At last, the loss function becomes equation \ref{equation1} as mentioned at the beginning. Now our task is to maximize this log likelihood. As we can see, when we are maximizing the log likelihood, the derivative of the score function is:
\begin{equation}
e_1^{t+1}=e_2^{t} + e_1^{t} \cdot diag (f(x_{ij}^{t}))
\end{equation}
\begin{equation}
e_2^{t+1}=e_1^{t} + e_2^{t} \cdot diag (f(x_{ij}^{t}))
\end{equation}
It can be seen if $e_1$ and $e_2$ share same latent pattern of the expression $x_{ij}$, they will enhance this pattern when we are maximizing the loss function. Thus expressions which are highly correlated with the same relation will be grouped together in the vector space. As mentioned in \ref{expression}, we assume the expressions are highly correlated with relations.\\
\paragraph{Word Prediction Loss Function}
The Word Prediction Loss Function $L_i$ is at internal nodes. Its formula is the same as Skip-gram's:\\
\begin{equation}
	\log\ \sigma(v_c\cdot v_w) + \sum_{i=1}^{k}E_{w_i \sim P_n(w)}[\log\ \sigma(-v_{c'_i} \cdot v_w)]
\end{equation}
The difference is that $v_w$ is not the word representation now, it is the phrase representation computed by internal node of LSTM. And $v_c$ are the context words of this phrase. For example, in figure \ref{pathtotree}, $v_c$ for internal node which outputs the representation of "Stanford Research" is word vector of "at" if the context window is one. We can see this loss function is for measuring the phrase representation's correlation with its local contexts.\\
Although local contexts around phrases don't have correlations with relations as strong as entities, they still include some semantic information of relations. Loss function $L_o$ at output node implies that the expressions which have same entities around have potential to own the same relation. Here we want to add more implications. The combination of entity prediction loss $L_o$ at output node and word prediction loss $L_i$ at internal node implies that the expressions which have the same entities  and local contexts around have potential to own the same relation. 
\paragraph{Options for Parameters Learning}
\label{updateweights}
After loss calculated, there are several choices when learning weights. We can choose to only update the model parameters $W_l$ of the LSTM. We can also update the word vectors $v_w$ and entity vectors $v_e$ in the repositories $E_{w}$ and $E_{e}$. All the options will bring different performance, so we will test some possible variations to see which one will bring the highest performance.
\subsubsection{Fine-tuning}
Since pre-training is a greedy strategy, to fit our global interest--classify relations, there is still one more step, fine-tuning. As in figure \ref{fintuning}, we input both hand-crafted features and the expression representations mapped by LSTM to the logistic regression. During training, the model parameters of both LSTM $W_l$ and logistic regression $W_r$ will be learned with the logistic loss function. After training, we will get a new model which will bring different performance compared with two baseline models.\\
\chapter{Experiments}
\label{cha:Methodology}
In this chapter, we will introduce the details of our dataset and how we pre-process our dataset. Then the methods of hyperparameter tuning will be proposed along with the tuning results. After that, we will explain our experiment settings. At last, the experiment results and analysis will be displayed.
\section{Dataset Description}
\begin{figure}[htpb]
    \centering
    \includegraphics[width=\textwidth=6cm,height=8cm]{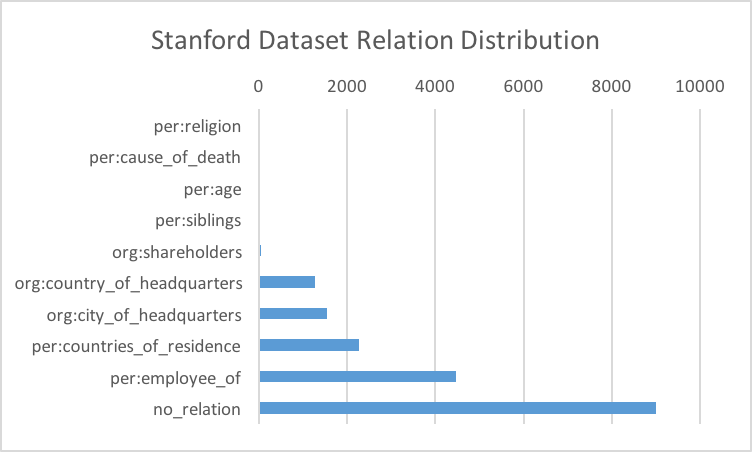}
    \caption{
        Distribution of Relations in Stanford Dataset.
    }
    \label{relationdistribution}
\end{figure}

\begin{figure}[htpb]
    \centering
    \includegraphics[width=\textwidth=6cm,height=8cm]{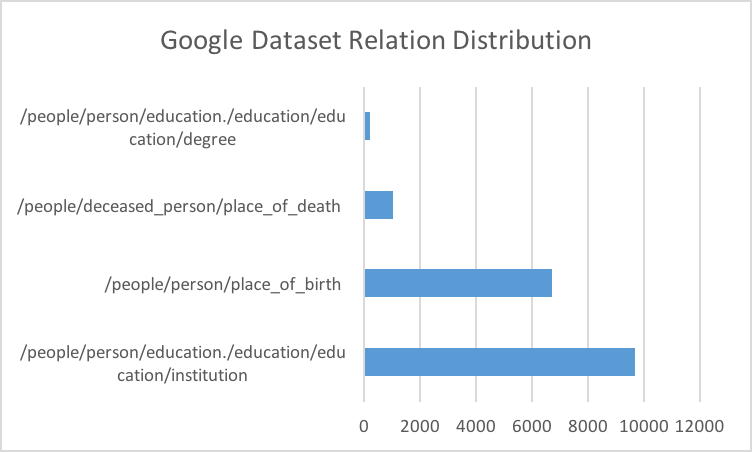}
    \caption{
        Distribution of Relations in Google Dataset.
    }
    \label{googlerelationdistribution}
\end{figure}

In our project, we use two datasets. Stanford Dataset, cited from \cite{2014emnlp-kbpactivelearning}, is from 2010 and 2013 KBP official document collections and a July 2013 dump of Wikipedia. Google dataset is extracted from Wikipedia, as introduced in Google Research Blog\footnote{http://googleresearch.blogspot.com.au/2013/04/50000-lessons-on-how-to-read-relation.html}.\\
\textbf{Stanford Dataset} is a CSV file. Each line contains one sentence, mentions, position index of mentions and relation label among mentions. We define one sentence as a data instance. Then through statistics, there are 28516 data instances and 40 relations. The distribution of relations is as in the figure \ref{relationdistribution}. We select the relations with the top five and the last five numbers, displayed in the figure. It can be seen the relations are not evenly distributed. The one with largest number is "\textit{no\_ relation}", which exists in 9002 instances while the one with the smallest number is "\textit{per:religion}" which only exists in 3 instances. \\
\textbf{Google Dataset} is a JSON file. Each line is a JSON object which contains one sentence, the machine id of the entities and the relation among two entities. There are 17582 instances in this dataset. There are only four relations. Relation distribution is as follows Section \ref{googlerelationdistribution}.
\subsection{Pre-processing Steps}
In our project, we need to annotate the Part-of-Speech tag of every token. And we also need to build a dependency tree for every sentence. The most important thing is that we need the entities and their positions in the sentence .\\
We use \textbf{Stanford Parser}\footnote{http://nlp.stanford.edu/software/lex-parser.shtml} to tokenize sentence and annotate Part-of-Speech for tokens. Then we use linear \textbf{MaltParser}\footnote{https://github.com/dkpro/dkpro-core/tree/bf90b7952c6efb3945a23ad21c04c57465839cb1} to build the dependency tree.\\
And since Stanford dataset only offers us mentions and their positions in sentence. We do the entity linkage for Stanford dataset. We link mentions to Freebase entities. Because there is no tool directly linking mentions to Freebase entities. First we link mentions to YAGO entities using AIDA.  AIDA is a framework and online tool for entity detection and disambiguation\footnote{https://www.mpi-inf.mpg.de/departments/databases-and-information-systems/research/yago-naga/aida/}. Then as introduced in \ref{freebaseyago}, both Freebase and YAGO are derived from Wikipedia, so we can map YAGO entities to Wikipedia entities using a mapping dictionary. Following that we dumped whole Freebase database, imported it to Apache Jena database and extract entities from Freebase based on the Wikipedia entity ids. Every entity in Freebase corresponds a unique machine id. So here we are able to use machine ids to symbolize every entity.\\
For Google dataset, it gives us the machine id of Freebase, so we use Freebase REST API to get the entity name. Since Google dataset doesn't give us the position of every entity, we use a heuristic regular expression matching to find their positions. There are three kinds of entities, the location, education degree and person name. For a person name, it is not quite possible to have ambiguation in a sentence. Thus if part of the name can be matched at the specific position in a sentence, we assume the name's entity is at that position. For the other kinds of entities, all the parts of their entity text should be exactly matched.
\subsection{Datasets Analysis}
The two datasets are not compatible. We will assign them different tasks.\\
Compared with Google dataset, there are several challenges for our model when dealing with Stanford data set. First, there are only 28516 data instances which is quite a small number. Second, there are 40 relations which are not evenly distributed. This might bring difficulties for our supervised learning model. Thirdly, the reasons for "\textit{no\_relation}" might vary but all the data instances with label "\textit{no\_relation}" are considered in the same group. Those instances would create noise. Fourth, the instances with "\textit{no\_relation}" occupy one third of whole dataset, which may lead a huge bias.\\
However the limitations of Google dataset is obvious as well. First, the data size is even smaller than Stanford dataset. Secondly, it has only four relations which makes this dataset not convincible enough to evaluate our model.\\
Thus here, we only use Google dataset for hyperparameter tuning during pre-training phase because less relations can make the tuning results easier to compare and the computation cost is lower. All the other tasks during pre-training phase and fine-tuning phase including training, validation and testing will use Stanford dataset. We believe the results acquired by Stanford dataset will be more sound.
\section{Hyperparameter Tuning}
In this section, the methods we used to do hyperparameter tuning will be introduced. And the results of tuning will be displayed.
\subsection{Hyperparameter Tuning for Pre-training}
We use Google dataset to do the hyperparameter tuning for pre-training.\\ 
We choose 3/4 Google dataset as training set and 1/4 as validation set. There is no test data because pre-training is unsupervised. It should be noted the dataset is not strictly randomly separated. We want to assure the validation set will have four relations as well as the training set. So we randomly split the dataset for many times. If there are no four relations in validation set, we re-split. After splitting, the partition of dataset won't be changed any more because we want, every time, the tuning results can be compared horizontally with other parameter settings.\\
We choose L1 regularization here. Although entity vectors and word vectors are not from the same domain, we still learn them with shared learning rates for reducing computational cost. And we use k-means to cluster the expression representations after training. After that, Rand Index introduced in \ref{evaluation} is used the evaluate the performance of the model.\\
We use different learning rates and regularization parameters for learning model parameters and updating input vectors.\\
We firstly tune the learning rates $\alpha_{pw}$ and regularization parameter $\lambda_{pw}$ for learning model parameters of LSTM. And we turn off the options of learning word vectors and entity vectors. Then we train the model and observe the results of Rand Index as table \ref{tab:machines}. We can see that different L1 regularization barely bring different effect while learning rates of 0.001f bring a large variance for the results. So we don't select 0.001f as learning rate. Instead we select the 0.001f and 0.1f as the optimal $\lambda_{pw}$ and $\alpha_{pw}$ respectively.\\
Then we tune the learning rates $\alpha_{pe}$ and regularization parameters $\lambda_{pe}$ for learning entity vectors and word vectors with the fixed tuned LSTM parameters $\alpha_{pw}$ and $\lambda_{pw}$. In comparison with the tuning results of LSTM, there is not much variance. All the results are around 0.58f. So we just choose a commonly adopted pair. Empirically, $\alpha_{pe}$ could be 10 times $\alpha_{pw}$. Thus we select 0.001f and 0.1f as the optimal regularization parameter $\lambda_{pe}$ and learning rate $\alpha_{pe}$ respectively.
\begin{table*}
  \centering
  \begin{adjustbox}{width=1\textwidth}
  
{
\sffamily

\begin{tabular}{r@{\hspace{2ex}}r@{\hspace{2ex}}r@{\hspace{2ex}}r@{\hspace{2ex}}r@{\hspace{2ex}}r@{\hspace{2ex}}r}
\\[-2ex]
{\textbf{Regularization Parameter}} & \multicolumn{1}{r@{\hspace{1.5ex}}}{\textbf{0.0001f}} &  \multicolumn{1}{r@{\hspace{1.5ex}}}{\textbf{0.0001f}} & \multicolumn{1}{r@{\hspace{1.5ex}}}{\textbf{0.0001f}} & \multicolumn{1}{r@{\hspace{1.5ex}}}{\textbf{0.000001f}} & \multicolumn{1}{r@{\hspace{1.5ex}}}{\textbf{0.000001f}} & \multicolumn{1}{r}{\textbf{0.000001f}}\\
{\textbf{Learning Rate}} & \multicolumn{1}{r@{\hspace{1.5ex}}}{\textbf{0.1f}} &  \multicolumn{1}{r@{\hspace{1.5ex}}}{\textbf{0.01f}} & \multicolumn{1}{r@{\hspace{1.5ex}}}{\textbf{0.001f}} & \multicolumn{1}{r@{\hspace{1.5ex}}}{\textbf{0.1f}} & \multicolumn{1}{r@{\hspace{1.5ex}}}{\textbf{0.01f}} & \multicolumn{1}{r}{\textbf{0.001f}}\\
\midrule 
{\textbf{Rand Index}} & 
0.5850123 & 0.5850123 & 0.5850129 & 0.5850123 & 0.5850123 & 	0.56168 \\
{\textbf{}} & 
0.5850123 & 0.5850123 & 0.58501357 & 0.5850123 & 0.5850123 & 0.58501357 \\
{\textbf{}} & 
& &	0.5850123& &	 &	0.5850123\\
{\textbf{}} &
& &	0.47688133& &	 &	0.47568637\\
{\textbf{}} &
& &	0.47455508 & &	 &	0.47704232\\
{\textbf{}} & 
& &	0.47480485& &	 &	\\

\bottomrule
\end{tabular}
}

  \end{adjustbox}
  \caption{Tuning Results for Weights on LSTM}
  \label{tab:machines}
\end{table*}
\subsection{Hyperparameter Tuning for Fine-tuning}
We use Stanford Dataset to do the hyperparameter tuning for fine-tuning. 35\% of the dataset, which is 9980 instances, is training set. 20\% of the dataset is both the validation set and test set.\\
In hyperparameter tuning, we don't use any pre-trained models, the model parameters of LSTM are randomly initialized.\\
Here we tune the learning rates $\alpha_{fw}$ and $\alpha_{fc}$, regularization parameters $\lambda_{fw}$ and $\lambda_{fc}$ for LSTM and logistic regression respectively and separately.\\
Firstly, we fix LSTM learning rate $\alpha_{fw}$ as 0.01f and regularization parameter $\lambda_{fw}$ as 0.001f. Then we design 6 pairs of $\lambda_{fc}$ and $\lambda_{fc}$. The results on the test set is as table \ref{tab:finetune1}. As we can see the pair of $\alpha_{fc}$ as 0.1f and $\lambda_{fc}$ as 0.000001f can bring the best recall, F1Measure and accuracy. Although precision is not the best, it is not far from the optimal performance. So we decide to choose $\alpha_{fc}$ as 0.1f and $\lambda_{fc}$ as 0.000001f.\\
\begin{table*}
  \centering
  \begin{adjustbox}{width=1\textwidth}
  
{
\sffamily
\begin{tabular}{r@{\hspace{1.5ex}}r@{\hspace{1.5ex}}r@{\hspace{1.5ex}}r@{\hspace{1.5ex}}r@{\hspace{1.5ex}}r@{\hspace{1.5ex}}r}
\\[-2ex]
{\textbf{Regularization Parameter}} & \multicolumn{1}{r@{\hspace{1.5ex}}}{\textbf{0.0001f}} &  \multicolumn{1}{r@{\hspace{1.5ex}}}{\textbf{0.0001f}} & \multicolumn{1}{r@{\hspace{1.5ex}}}{\textbf{0.0001f}} & \multicolumn{1}{r@{\hspace{1.5ex}}}{\textbf{0.000001f}} & \multicolumn{1}{r@{\hspace{1.5ex}}}{\textbf{0.000001f}} & \multicolumn{1}{r}{\textbf{0.000001f}}\\
{\textbf{Learning Rate}} & \multicolumn{1}{r@{\hspace{1.5ex}}}{\textbf{0.1f}} &  \multicolumn{1}{r@{\hspace{1.5ex}}}{\textbf{0.01f}} & \multicolumn{1}{r@{\hspace{1.5ex}}}{\textbf{0.001f}} & \multicolumn{1}{r@{\hspace{1.5ex}}}{\textbf{0.1f}} & \multicolumn{1}{r@{\hspace{1.5ex}}}{\textbf{0.01f}} & \multicolumn{1}{r}{\textbf{0.001f}}\\
\midrule 
{\textbf{Macro-Precision}} & 
0.4064905& 0.31999898 & 0.01934031 & 0.42622223 & 0.4421185 & 	0.044637516 \\
{\textbf{Macro-Recall}} & 
0.2545123 & 0.12995863 & 0.029275449 & 0.2593944 & 0.18361308 & 0.031583756 \\
{\textbf{Macro-F1Measure}} 
&0.29914954 & 0.14870085& 0.018138776 & 0.3049322	& 0.22467111&	0.0220251\\
{\textbf{Accuracy}} 
&0.5206872 &	0.4652875& 0.33134642 &	0.5322581 & 0.49964938 &	0.33520338\\
\bottomrule
\end{tabular}
}
  \end{adjustbox}
  \caption{Tuning Results for Model Parameters of Logistic Regression}
  
  \label{tab:finetune1}
\end{table*}
Secondly, we tune $\alpha_{fw}$ and $\lambda_{fw}$. We fix the $\alpha_{fc}$ as 0.1f and $\lambda_{fc}$ as 0.000001f. As well, we design 6 pairs. The result on the test set is as table \ref{tab:finetune2}. Then we found the pair of $\alpha_{fw}$ as 0.1f and $\lambda_{fw}$ as 0.000001f could bring the best accuracy. Although compared with other pairs, Precision, Recall and F1Measure are not the best, none of the performance is the worst. This pair is the most stable one. Thus we decide to choose $\alpha_{fw}$ as 0.1f and $\lambda_{fw}$ as 0.000001f.\\
\begin{table*}
  \centering
  \begin{adjustbox}{width=1\textwidth}
  
{
\sffamily
\begin{tabular}{r@{\hspace{1.5ex}}r@{\hspace{1.5ex}}r@{\hspace{1.5ex}}r@{\hspace{1.5ex}}r@{\hspace{1.5ex}}r@{\hspace{1.5ex}}r}
\\[-2ex]
{\textbf{Regularization Parameter}} & \multicolumn{1}{r@{\hspace{1.5ex}}}{\textbf{0.0001f}} &  \multicolumn{1}{r@{\hspace{1.5ex}}}{\textbf{0.0001f}} & \multicolumn{1}{r@{\hspace{1.5ex}}}{\textbf{0.0001f}} & \multicolumn{1}{r@{\hspace{1.5ex}}}{\textbf{0.000001f}} & \multicolumn{1}{r@{\hspace{1.5ex}}}{\textbf{0.000001f}} & \multicolumn{1}{r}{\textbf{0.000001f}}\\
{\textbf{Learning Rate}} & \multicolumn{1}{r@{\hspace{1.5ex}}}{\textbf{0.1f}} &  \multicolumn{1}{r@{\hspace{1.5ex}}}{\textbf{0.01f}} & \multicolumn{1}{r@{\hspace{1.5ex}}}{\textbf{0.001f}} & \multicolumn{1}{r@{\hspace{1.5ex}}}{\textbf{0.1f}} & \multicolumn{1}{r@{\hspace{1.5ex}}}{\textbf{0.01f}} & \multicolumn{1}{r}{\textbf{0.001f}}\\
\midrule 
{\textbf{Macro-Precision}} & 
0.4220812& 0.43920225 & 0.443421 & 0.4165563 & 0.4442537 & 	0.43949598 \\
{\textbf{Macro-Recall}} & 
0.24918497 & 0.25197038 & 0.25125498 & 0.25176826 & 0.24653225 & 0.25963798 \\
{\textbf{Macro-F1Measure}} 
&0.2968473 & 0.30054057& 0.2980951 & 0.2955865	& 0.29633698 &	0.30784795\\
{\textbf{Accuracy}} 
&0.5217391 &	0.5269986& 0.5245442 &	0.5336606 & 0.52384293 &	0.5269986\\
\bottomrule
\end{tabular}
}

  \end{adjustbox}
  \caption{Tuning Results for Model Parameters of LSTM}
  
  \label{tab:finetune2}
\end{table*}
\section{Experiment Settings}
Based on our model variations, we designed eight experiments, two for baseline and six for our model. For all the experiments, we choose 70\% of our dataset(19961 instances) as out training set, 10\%(2851 instances) as our validation set and 20\%(5704 instances) as our testing set. We adopt an incrementing instance training strategy. At beginning, we only use $2^{-9}$ of whole training instances(29 instances). Then we double the training instances every time until the number of instances comes to 19961. At last, there will be 10 times training for every experiment and we record each of the results.\\
There are invariants and variants among the experiments. The next we will introduce the variant and invariant settings.
\subsection{Baseline}
\subsubsection{Invariants}
For the two baselines, the learning rate $\alpha_{fc}$ and regularization parameter $\lambda_{fc}$ for logistic regression are 0.1f and 0.000001f respectively.\\
\subsubsection{Variants}
The variants for two baselines are as follows:\\
\begin{enumerate}[I.]
  \item For the first baseline, we only use hand-crafted feature. We refer it as $B-H$ (Baseline with Hand-crafted Feature).
  \item For the second baseline, we use both expression representation and handcrafted feature. However unlike the pre-trained model , the model parameters of LSTM here is random initialized. We refer it as $B-HL$ (Baseline with Hand-crafted feature and Shortest Dependency Path LSTM ).
\end{enumerate}
\subsection{Representation Learning Models}
\subsubsection{Invariants}
In pre-training phase, we select the 0.1f and 0.001f as the optimal LSTM learning rate $\alpha_{pw}$ and LSTM regularization parameter $\lambda_{pw}$ respectively. And for updating entity and word vectors, we select 0.1f and 0.001f as the optimal learning rate $\alpha_{pe}$ and regularization parameter $\lambda_{pe}$ respectively.\\
In fine-tuning phase, we select learning rate $\alpha_{fc}$ as 0.1f and regularization parameter $\lambda_{fc}$ as 0.000001f for logistic regression. And we choose learning rate $\alpha_{fw}$ as 0.1f and regularization parameter $\lambda_{fw}$ as 0.000001f for LSTM. For the negative sampling, we set the sample times as 5 for both the entity prediction loss $L_o$ and word prediction loss $L_i$ .\\
\subsubsection{Variants}
The experiments can be categorized by whether using word prediction loss $L_i$ or not during pre-training phase.\\
The experiments without word prediction loss $L_i$ are as follows:\\
\begin{enumerate}[I.]
  \item We set this experiment as only updating model parameter of LSTM in pre-training phase. We don't use word prediction loss, don't update entity vectors and word vectors. We refer this experiment as $M-E-L$ (Model with Entity Prediction Loss, Shortest Dependency Path LSTM).
  \item In this experiment, we don't use word prediction loss, don't update word vectors but update entity vectors. We refer this experiment as $M-E-LUE$ (Model with Entity Prediction Loss, Shortest Dependency Path LSTM, Update Entity Vector).
  \item In this experiment, we don't use word prediction loss but update word vectors and entity vectors. We refer this experiment as $M-E-LUEW$ (Model with Entity Prediction Loss, Shortest Dependency Path LSTM, Update Entity Vector and Word Vector).
\end{enumerate}
The experiments with word prediction loss $L_i$ are as follows:\\
\begin{enumerate}[I.]
  \item We set this experiment as using word prediction loss. We don't update entity vectors and word vectors. We refer this experiment as $M-EW-L$ (Model with Entity Prediction Loss and Word Prediction Loss, Shortest Dependency Path LSTM).
  \item In this experiment, we use word prediction loss, don't update word vectors but update entity vectors. We refer this experiment as $M-EW-LUE$ (Model with Entity Prediction Loss and Word Prediction Loss, Shortest Dependency Path LSTM,Update Entity Vector).
  \item In this experiment, we use word prediction loss, update word vectors and entity vectors. We refer this experiment as $M-EW-LUEW$ (Model with Entity Prediction Loss and Word Prediction Loss, Shortest Dependency Path LSTM,Update Entity Vector, Word Vector).
\end{enumerate}
\section{Results}
\label{cha:result}
In this section, we will present the results and analysis of our experiments. We will focus on the comparison between the text representation models and the two baselines.
\subsection{Overall Comparison}
\label{overallcompare}
\begin{figure*}
  \label{fig:ap}
  \subfigure[Performance of $A-B$\label{fig:ab}]{\includegraphics[width=0.5\columnwidth]{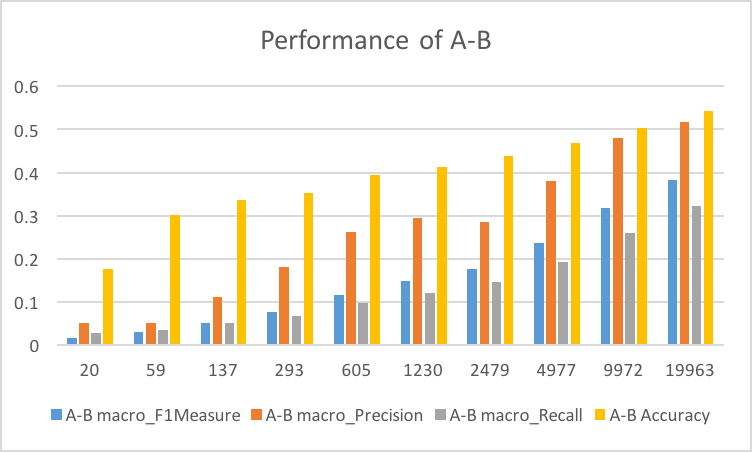}}
  \subfigure[Performance of $A-M$\label{fig:am}]{\includegraphics[width=0.5\columnwidth]{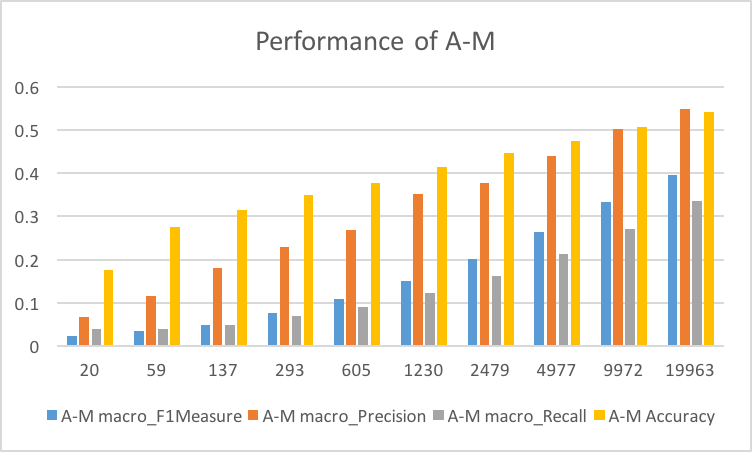}}
  \caption{Performance of the Baselines and Models based on Representation Learning}
\end{figure*}

\begin{figure}[htpb]
    \centering
    \includegraphics[width=1.0\columnwidth]{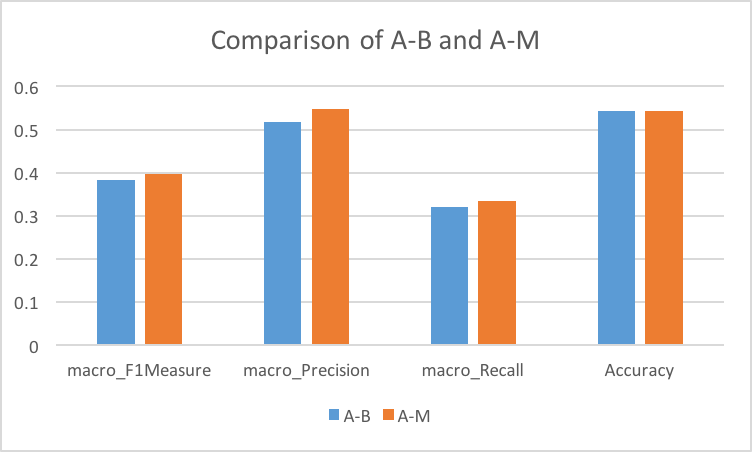}
    \caption{
        Comparison of the $A-B$ and $A-M$ when number of training instances is 19963.
    }
    \label{a:averageperformance}
\end{figure}

\begin{figure}[htpb]
    \centering
    \includegraphics[width=1.0\columnwidth]{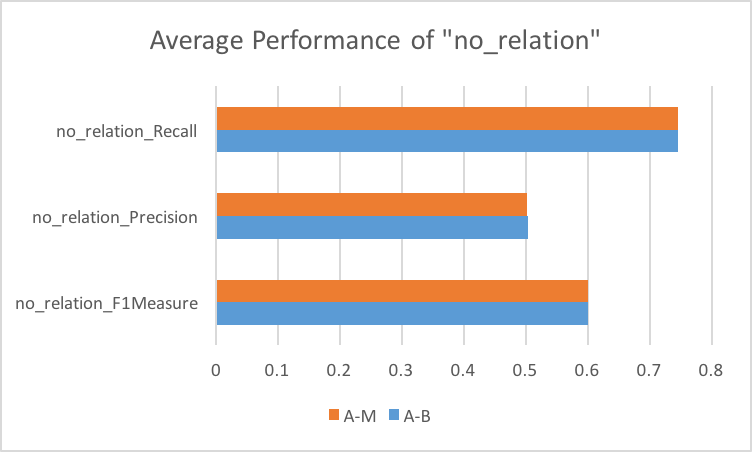}
    \caption{
        Comparison of the average performance $A-B$ and $A-M$ on the relation \textit{"no\_ relation"}.
    }
    \label{norelation}
\end{figure}

We compared the average performance $A-B$ of the two baselines and the average performance $A-M$ of the six text representations models. From chart \ref{fig:ap}, we can see that both of the performance $A-B$ and $A-M$ are increasing with the increasing number of training instances. It proves that although different models will bring different performance relatively, number of training instances is still the dominant factor influencing the model performance. And we also realized that when the size of training instances is small, the model performance is usually pretty lousy. Under this situation, the performance is also unstable. Comparing the performance between models on a small instance size is meaningless. Thus we only compare the performance with largest number of instances (19963) in the next.\\
Let's take a close look at the case with 19963 training instances. From the chart \ref{a:averageperformance}, the macro-precision of $A-M$ is improved from 0.51720151 to 0.548091515, macro-recall is improved from 0.321693825 to 0.33483990333 and macro-FMeasure is improved from 0.382639875 to 0.396789315. However, there is not much change on accuracy. The accuracy of $A-B$ is 0.5435659 and the accuracy of $A-M$ is 0.54260168333.\\
As results shown, the pre-training improved the performance of baseline w.r.t all the macro performances. However accuracy doesn't look much different. Because the performance of accuracy is mainly dominated by the relations with large instance size while macro performance consider all the relations equal. So we can infer, with this dataset, pre-training doesn't improve performance on the relations with large instance size but it is able to help logistic regression with detecting the relations with small instance size.\\
We found in the dataset, there are 9002 instances with relation \textit{no\_ relation}, which occupies 1/3 of the whole dataset. So we compare the $A-B$ to $A-M$ on the relation \textit{no\_ relation}. From the chart \ref{norelation}, we can see that $A-B$ and $A-M$ on the relation \textit{no\_ relation} are almost the same. This can explain why the accuracy isn't improved. Because pre-training has not much effect on \textit{no\_ relation} while \textit{no\_ relation} dominates the performance of accuracy. \\
There are many possible reasons that pre-training doesn't influence the performance on \textit{no\_ relation}. We conjecture it is because \textit{no\_ relation} doesn't really mean there is no relation between two mentions. It just means there is no relation in the label set. For example, here is a sentence in dataset: \textit{Shortly after this; the then Chancellor of the Exchequer; Nigel Lawson; and other Conservative politicians claimed that misleading statistics were largely responsible for the Government's poor handling of the economy .} The mentions in this sentence are \textit{Nigel Lawson} and \textit{Exchequer}. We can see there is a relation \textit{ChancellorOf} between two mentions although the instance is still labelled as \textit{no\_ relation}. In this case, our model may still find the pattern of \textit{ChancellorOf} and highlight it. However this is a misleading information for logistic regression because logistic regression will only try to detect the relations which are pre-defined by human.\\
However pre-training has a quite good performance on relations with a small number. For example, the relation \textit{per:cause\_ of\_ death} only has 5 instances. In both baselines, the precision, recall and FMeasure on this relation are all 0 when training instances are with the maximum size(19963). In the six experiments of new models, there are three of them ($M-E-LUE$, $M-E-LUEW$, $M-EW-LUEW$) which is capable of detecting this relation . And precision, recall and FMeasure are all 1. Because the performance on relations with small number increase and macro consider every relation equal weight, the whole performance of precision, recall and FMeasure of $A-M$ are increased compared with $A-B$.
\subsection{Evaluation of Baseline}
\begin{figure*}
  \label{fig:baseline}
  \subfigure[Performance of $B-H$\label{fig:baseline1}]{\includegraphics[width=0.5\columnwidth]{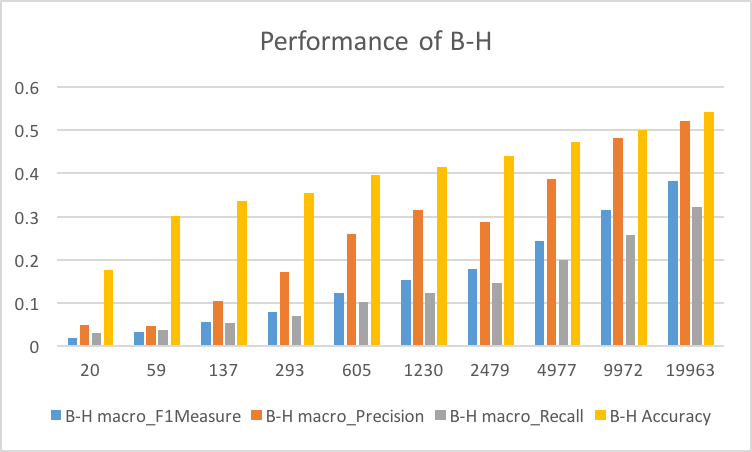}}
  \subfigure[Performance of $B-HL$\label{fig:baseline2}]{\includegraphics[width=0.5\columnwidth]{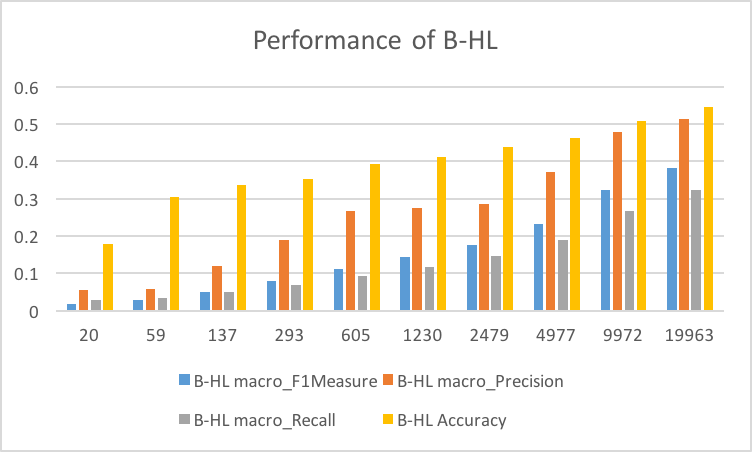}}
  \caption{Performance of the Baselines}
\end{figure*}

From chart \ref{fig:baseline}, we can see that although baseline $B-HL$ imported the expression representation, the performances of the two baselines are almost the same with respect to macro-precision, macro-recall, macro-FMeasure and accuracy. The difference between them is that the precision of $B-HL$ is always increasing with the increasing number of training instances while precision of $B-H$ decreased when number of training instances is 2479. The reason might be the result is unstable when experimenting with small number of training instance. We can see after the  training instance size becomes larger, the growth curve of $B-H$ becomes smooth. So we can infer that expression representation learned with random initialization LSTM has no either positive or negative influence on the baseline system with hand-crafted feature.
\subsection{Influence of Entity Prediction Loss}
In order to observe the influence of global loss, we compare the performance of $M-E-L$ with the performance of two baselines $B-H$, $B-HL$ and the average performance $A-M$ of representation learning models.\\
It can be seen that with respect to precision, recall and FMeasure,  the performance of $M-E-L$ is higher than performance of two baselines. This can display that global loss function has the ability to help detect the relation. But in comparison with $A-M$, the performance of $M-E-L$ with respect to precision, recall and FMeasure are all a little lower. $M-E-L$ doesn't achieve the average level, which infers that the reason that $A-M$ performs better is depended on variations other than Entity Prediction Loss.
\begin{figure}[htpb]
    \centering
    \includegraphics[width=1.0\columnwidth]{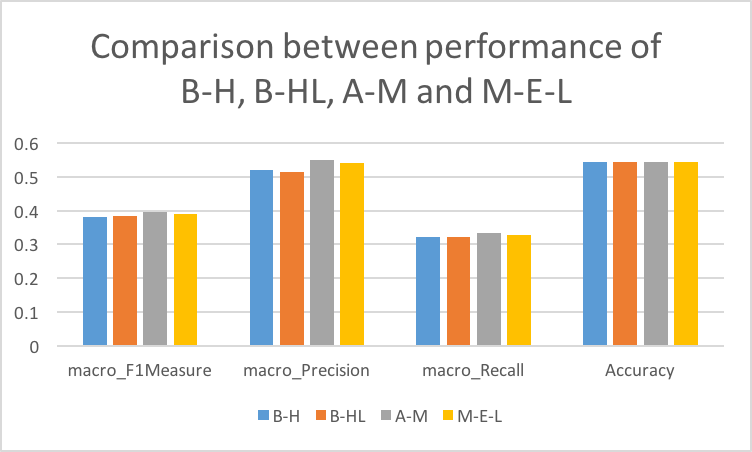}
    \caption{
        Compare the performance of $B-H$, $B-HL$, $A-M$ and $M-E-L$.
    }
    \label{b:averageperformance}
\end{figure} 
\subsection{Influence of Word Prediction Loss}
In order to observe the influence of word prediction loss, we compare the performance of $M-EW-L$ with the performance of $M-E-L$ and the average performance $A-M$ of our novel model with six variations.\\
It can be seen after we imported the word prediction loss, precision has slightly improvement compared with $M-E-L$. The others are all decreasing. This displays that if we only update model parameters, the word prediction loss may not be helpful on improving performance. However we can observe whether word prediction loss combine with others will bring different outcomes.
\begin{figure}[htpb]
    \centering
    \includegraphics[width=1.0\columnwidth]{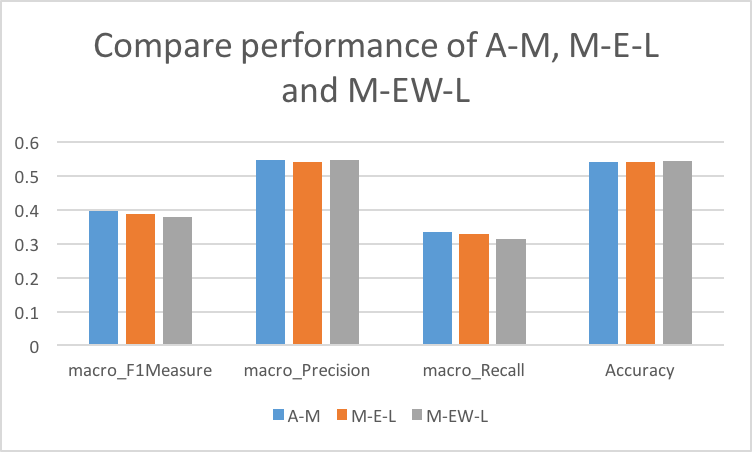}
    \caption{
        Compare the performance of $A-M$, $M-E-L$ and $M-EW-L$.
    }
    \label{averageperformance}
\end{figure} 
\subsection{Influence of Updating Entity Vector}
To observe the influence of updating entity vectors, we separate the comparison as two pairs. One pair is the comparison between $M-E-L$, $M-E-LUE$ and $A-M$. Another pair is the comparison between $M-EW-L$, $M-EW-LUE$ and $A-M$. We want to observe the results in the conditions with and without word prediction loss.\\
\begin{figure*}
  \label{fig:baseline_mel}
  \subfigure[Compare performance of $M-E-L$, $M-E-LUE$ and $A-M$\label{fig:e1}]{\includegraphics[width=0.5\columnwidth]{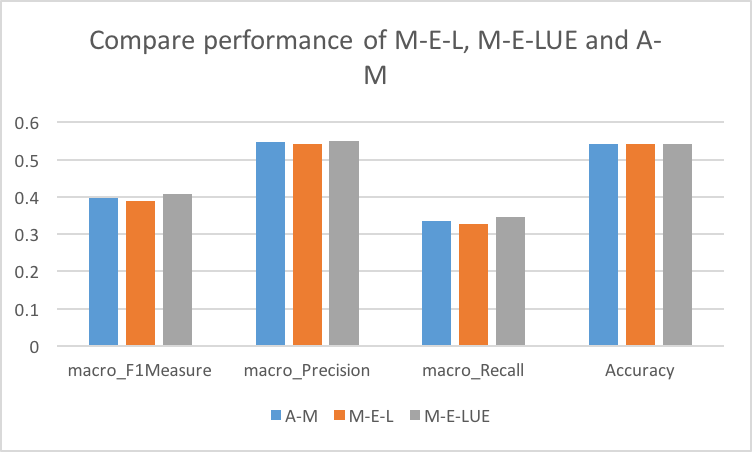}}
  \subfigure[Compare performance of $M-EW-L$, $M-EW-LUE$ and $A-M$\label{fig:e2}]{\includegraphics[width=0.5\columnwidth]{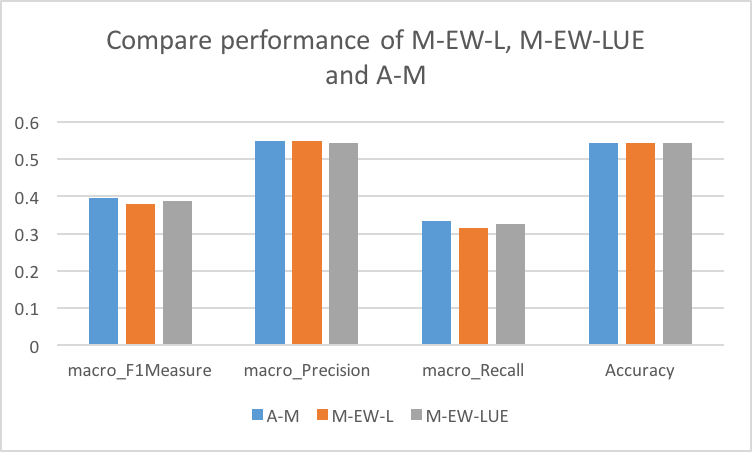}}
  \caption{Performance of the Updating Entity Vector}
\end{figure*}
From the chart \ref{fig:e1}, we can see that, without word prediction loss, after updating the entity vectors, all the performance except accuracy is increased and better than average performance. Precision is increased from 0.5408333 to 0.5501761. And 0.5501761 is the highest precision we've seen for now. Recall is increased from 0.32829905 to 0.34533334. FMeasure is increased from 0.38919714 to 0.4070243.\\
However from chart \ref{fig:e2}, we can see that, with word prediction loss and updating entity vectors, not all the performances are increased. The precision is decreased from 0.54792714 to 0.543988. Compared with the corresponding situation without word prediction loss, word prediction loss doesn't have a good performance here.\\
Here is the conjectural reason. Updating entity vector can help pre-training and group the expression representations which have the same entities around. However word prediction loss has no relation with entities prediction. Although $M-EW-LUE$ also use the entity prediction loss, the word prediction loss may have some negative influence on updating entity vectors. Thus the precision in $M-EW-LUE$ is not increased.
\subsection{Influence of Updating Word Vector}
To observe the influence of updating word vectors, we separate the comparison as two pairs as well. One pair is the comparison between $M-E-LUE$, $M-E-LUEW$ and $A-M$. Another pair is the comparison between $M-EW-LUE$, $M-EW-LUEW$ and $A-M$. We want to observe the results in the conditions with and without word prediction loss.\\
\begin{figure*}
  \label{fig:baseline_am}
  \subfigure[Compare performance of $A-M$, $M-E-LUE$ and $M-E-LUEW$\label{fig:c1}]{\includegraphics[width=0.5\columnwidth]{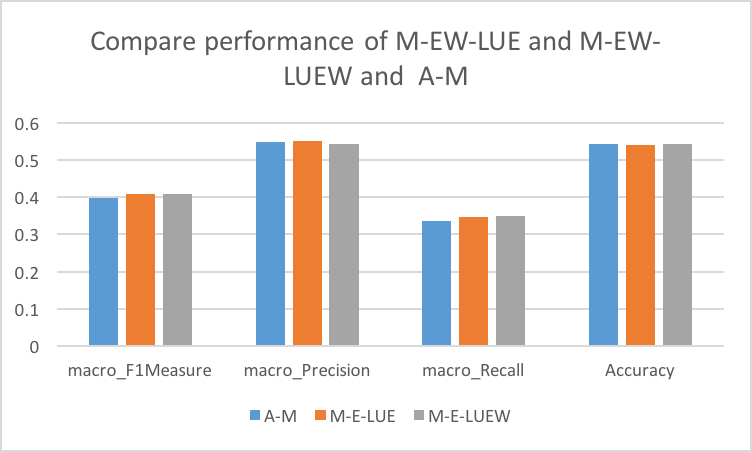}}
  \subfigure[Compare performance of $A-M$, $M-EW-LUE$ and $M-EW-LUEW$\label{fig:c2}]{\includegraphics[width=0.5\columnwidth]{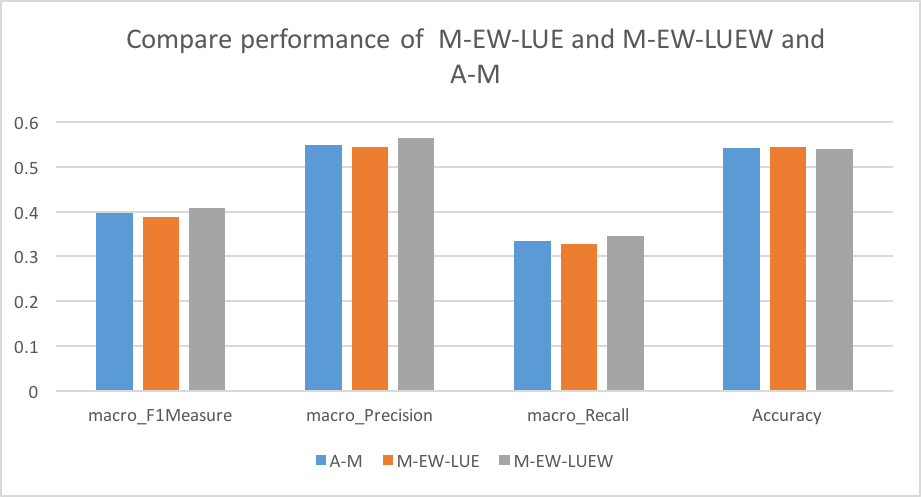}}
  \caption{Performance of the Updating Word Vector}
\end{figure*}
From chart \ref{fig:c1}, we can see that from $M-E-LUE$ to $M-E-LUEW$, the model's precision is decreased from 0.5501761 to 0.5420345. The recall, FMeasure and Accuracy are almost the same. We added a new variation but performance is not improved. The reason might be the same as that we add word prediction loss when updating entity vectors. Updating word vector alone without word prediction loss has a negative influence on the models with updating entity vectors and entity prediction loss.\\
However, from chart \ref{fig:c2}, we can see that we get the highest performance for now. Precision is increased from 0.543988 to 0.56359005, recall is increased from 0.38833365 to 0.40700668 and FMeasure is increased from 0.3269379 to 0.34504116.\\
From the observation, I think the reasons might be as follows:\\
Let's denote the variations with two pairs, word prediction loss and updating word vectors, entity prediction loss and updating entity vectors. Updating entity vector can help pre-training with entity prediction loss and updating word vector can help pre-training with word prediction loss. However, when they exist alone, none of them will achieve the same performance as they exist in pair. And when the two pairs all exist, like the situation in $M-EW-LUEW$, the performance of our model will be improved to the largest degree.
\subsection{Summary}
In sum, with 8 experiments, we can see that text representation learning has a positive influence on relation extraction task. With some specific combinations of variations, we can adjust the performance of models to the optimal. But it also has some limitations, especially when classifying the \textit{no\_ relation}.
\chapter{Conclusion}
This chapter will recap the contributions of our work and bring forward the future directions.
\section{Contributions} 
In this thesis, we presented our ideas about how to apply text representation learning on the relation extraction task. We also built several text representation learning models and conducted a series of experiments. From the results of experiments, we've proved that text representation is able to improve the performance of baseline system. And we also analysed the influence of every variation in our models. At last, we found an optimal combination of variations on our dataset.
\section{Future Work}
\label{sec:future}
From the results of experiments, we can see There is still much work ahead. In what follows we discuss several directions in future work.
\subsection{Pre-training on Different Dataset}
From our assumption in \ref{overallcompare}, the reason the performance of accuracy is not improved is that there are too many \textit{no\_ relation} instances. To justify our assumption and find the method to improve performance of accuracy, we could test our models on another dataset which have close numbers of relations as Stanford Dataset but without \textit{no\_ relation}. Then we could observe the performance and see if our assumption is right.\\
\subsection{Choice of Expression}
In our project, we build LSTM tree with the shortest dependency path between two mentions. The advantage is that the computational cost is relatively low. The disadvantage is that we lost some words, which might be important information. Thus we could choose to use the whole sequence between two mentions instead of just using the shortest path. Then we could test our models on Stanford Dataset again and observe the results.

\newpage
\appendix
\chapter{Appendix A}
\section{Results of Experiments} \label{App:AppendixA}
\begin{table*}
  \centering
  \begin{adjustbox}{width=1\textwidth, height=0.5\textheight}
  
{
\sffamily
\begin{tabular}{|c|c|c|c|c|c|c|c|c|c|c|}
	\hline
	instance\_number&20&59&137&293&605&1230&2479&4977&9972&19963\\
	\hline
	macro\_F1Measure&0.018139772&0.031910427&0.05459837&0.078539014&0.12296783&0.15323138&0.17747343&0.24234925&0.31485575&0.38220945\\
	\hline
	macro\_Precision&0.04787253&0.046931658&0.1032367&0.1718671&0.2588028&0.3142514&0.2868303&0.38719153&0.4817804&0.52024716\\
	\hline
	macro\_Recall&0.030014262&0.03794201&0.05397829&0.06860159&0.10219344&0.12313305&0.14694107&0.19825585&0.25670385&0.3213864\\
	\hline
	micro\_F1Measure&0.17724404&0.30014026&0.3367812&0.3536115&0.3951613&0.4156732&0.44004208&0.47230014&0.5012272&0.54242635\\
	\hline
	micro\_Precision&0.17724404&0.30014026&0.3367812&0.3536115&0.3951613&0.41567323&0.44004208&0.47230014&0.5012272&0.54242635\\
	\hline
	micro\_Recall&0.17724404&0.30014026&0.3367812&0.3536115&0.3951613&0.41567323&0.44004208&0.47230014&0.5012272&0.54242635\\
	\hline
	no\_relation\_F1Measure&0.21653256&0.47426277&0.5027556&0.50359964&0.5235507&0.54126036&0.55653507&0.5689549&0.5822034&0.59991145\\
\hline
	
\end{tabular}

}

  \end{adjustbox}
  \caption{Performance of $B-H$}
\end{table*}
\begin{table*}
  \centering
  \begin{adjustbox}{width=1\textwidth, height=0.5\textheight}
  
{
\sffamily
\begin{tabular}{|c|c|c|c|c|c|c|c|c|c|c|}
	\hline
	instance\_number&20&59&137&293&605&1230&2479&4977&9972&19963\\
	\hline
	macro\_F1Measure&0.017471176&0.028100075&0.05021503&0.07803785&0.11215027&0.14488265&0.176058&0.23334552&0.3224984&0.3830703\\
	\hline
	macro\_Precision&0.055989683&0.057870142&0.11891381&0.18958606&0.26699895&0.2743268&0.2852596&0.37197295&0.47980806&0.51415586\\
	\hline
	macro\_Recall&0.028256841&0.034659103&0.050094873&0.06808167&0.09242147&0.11720539&0.14570592&0.19010663&0.2657454&0.32200125\\
	\hline
	micro\_F1Measure&0.17829593&0.30452314&0.33590463&0.35396212&0.39323282&0.41146567&0.43723702&0.46353436&0.5078892&0.54470545\\
	\hline
	micro\_Precision&0.17829593&0.30452314&0.33590463&0.35396212&0.39323282&0.41146564&0.43723702&0.46353436&0.5078892&0.54470545\\
	\hline
	micro\_Recall&0.17829593&0.30452314&0.33590463&0.35396212&0.39323282&0.41146564&0.43723702&0.46353436&0.5078892&0.54470545\\
	\hline
	no\_relation\_F1Measure&0.22947441&0.4763508&0.49713394&0.506424&0.52562326&0.5375093&0.5530042&0.56059694&0.58682114&0.60229784\\
\hline
\end{tabular}

}

  \end{adjustbox}
  \caption{Performance of $B-HL$}
\end{table*}
\begin{table*}
  \centering
  \begin{adjustbox}{width=1\textwidth, height=0.5\textheight}
  
{
\sffamily
\begin{tabular}{|c|c|c|c|c|c|c|c|c|c|c|}
	\hline
	instance\_number&20&59&137&293&605&1230&2479&4977&9972&19963\\
	\hline
	macro\_F1Measure&0.020148417&0.035612766&0.045461733&0.07744566&0.109960504&0.15156762&0.20946942&0.26062033&0.33694676&0.38919714\\
	\hline
	macro\_Precision&0.045035914&0.114397734&0.1908015&0.20307654&0.25434348&0.34552377&0.39411435&0.44794875&0.5032449&0.5408333\\
	\hline
	macro\_Recall&0.029408913&0.03923494&0.047125094&0.07102449&0.092138685&0.12272665&0.16749257&0.20764077&0.27619955&0.32829905\\
	\hline
	micro\_F1Measure&0.16129032&0.27577138&0.32030153&0.34975457&0.375&0.41164097&0.45196354&0.47159886&0.5091164&0.5429523\\
	\hline
	micro\_Precision&0.16129032&0.27577138&0.32030153&0.34975457&0.375&0.41164094&0.45196354&0.47159886&0.5091164&0.5429523\\
	\hline
	micro\_Recall&0.16129032&0.27577138&0.32030153&0.34975457&0.375&0.41164094&0.45196354&0.47159886&0.5091164&0.5429523\\
	\hline
	no\_relation\_F1Measure&0.23804748&0.42401573&0.4806079&0.50677794&0.5158064&0.5390915&0.5631289&0.56986856&0.5838349&0.59854656\\
\hline
\end{tabular}

}

  \end{adjustbox}
  \caption{Performance of $M-E-L$}
\end{table*}
\begin{table*}
  \centering
  \begin{adjustbox}{width=1\textwidth, height=0.5\textheight}
  
{
\sffamily

\begin{tabular}{|c|c|c|c|c|c|c|c|c|c|c|}
	\hline
	instance\_number&20&59&137&293&605&1230&2479&4977&9972&19963\\
	\hline
	macro\_F1Measure&0.024447655&0.035612766&0.05191559&0.07818313&0.10998873&0.15913217&0.19980255&0.2834333&0.3414742&0.4070243\\
	\hline
	macro\_Precision&0.07726521&0.114397734&0.18967216&0.24354619&0.24314995&0.37582195&0.35995662&0.45521837&0.5111061&0.5501761\\
	\hline
	macro\_Recall&0.032938607&0.03923494&0.05249303&0.0694816&0.091346435&0.12667283&0.15975504&0.23382401&0.2766959&0.34533334\\
	\hline
	micro\_F1Measure&0.3108345&0.27577138&0.3096073&0.35220897&0.37885696&0.41602385&0.4482819&0.47685835&0.50508416&0.54102385\\
	\hline
	micro\_Precision&0.3108345&0.27577138&0.3096073&0.35220897&0.37885696&0.41602385&0.4482819&0.47685835&0.50508416&0.54102385\\
	\hline
	micro\_Recall&0.3108345&0.27577138&0.3096073&0.35220897&0.37885696&0.41602385&0.4482819&0.47685835&0.50508416&0.54102385\\
	\hline
	no\_relation\_F1Measure&0.48024145&0.42401573&0.47229174&0.50202715&0.52461857&0.5424103&0.5626315&0.57241523&0.5775333&0.5986696\\
\hline
\end{tabular}

}

  \end{adjustbox}
  \caption{Performance of $M-E-LUE$}
\end{table*}
\begin{table*}
  \centering
  \begin{adjustbox}{width=1\textwidth, height=0.5\textheight}
  
{
\sffamily
\begin{tabular}{|c|c|c|c|c|c|c|c|c|c|c|}
	\hline
	instance\_number&20&59&137&293&605&1230&2479&4977&9972&19963\\
	\hline
	macro\_F1Measure&0.029712686&0.035612766&0.0433363&0.074671&0.10396158&0.13905409&0.19994201&0.25954375&0.3273146&0.40842772\\
	\hline
	macro\_Precision&0.10237411&0.114397734&0.16286832&0.22117816&0.2632771&0.31922948&0.3558912&0.432315&0.49263152&0.5420345\\
	\hline
	macro\_Recall&0.035415888&0.03923494&0.04560056&0.06715876&0.0872334&0.11430035&0.16048174&0.20821652&0.26636833&0.34785005\\
	\hline
	micro\_F1Measure&0.26069424&0.27577138&0.31837308&0.35080644&0.37517536&0.41532254&0.4456522&0.4763324&0.50946707&0.54347825\\
	\hline
	micro\_Precision&0.26069424&0.27577138&0.31837308&0.35080644&0.37517533&0.41532257&0.4456522&0.4763324&0.50946707&0.54347825\\
	\hline
	micro\_Recall&0.26069424&0.27577138&0.31837308&0.35080644&0.37517533&0.41532257&0.4456522&0.4763324&0.50946707&0.54347825\\
	\hline
	no\_relation\_F1Measure&0.44153225&0.42401573&0.4802761&0.50145376&0.52176905&0.54518735&0.5573958&0.573401&0.58664984&0.6027821\\
\hline
\end{tabular}

}

  \end{adjustbox}
  \caption{Performance of $M-E-LUEW$}
\end{table*}
\begin{table*}
  \centering
  \begin{adjustbox}{width=1\textwidth, height=0.5\textheight}
  
{
\sffamily

\begin{tabular}{|c|c|c|c|c|c|c|c|c|c|c|}
	\hline
	instance\_number&20&59&137&293&605&1230&2479&4977&9972&19963\\
	\hline
	macro\_F1Measure&0.02462885&0.035612766&0.047480337&0.07955772&0.10714161&0.1608003&0.19433147&0.2571875&0.33160084&0.3807464\\
	\hline
	macro\_Precision&0.059920024&0.114397734&0.1951538&0.23618153&0.27573457&0.37409383&0.39589795&0.43779448&0.49749488&0.54792714\\
	\hline
	macro\_Recall&0.041950483&0.03923494&0.04750006&0.07140385&0.08962873&0.128514&0.15537168&0.2063707&0.2701213&0.31557792\\
	\hline
	micro\_F1Measure&0.113604486&0.27577138&0.31258765&0.34765077&0.37780505&0.41830295&0.44477558&0.47457924&0.5077139&0.5438289\\
	\hline
	micro\_Precision&0.113604486&0.27577138&0.31258765&0.34765077&0.37780505&0.41830295&0.44477558&0.47457924&0.5077139&0.5438289\\
	\hline
	micro\_Recall&0.113604486&0.27577138&0.31258765&0.34765077&0.37780505&0.41830295&0.44477558&0.47457924&0.5077139&0.5438289\\
	\hline
	no\_relation\_F1Measure&0.17611942&0.42401573&0.46968907&0.49907917&0.5244732&0.548093&0.56067264&0.5686315&0.585511&0.601929\\
\hline
\end{tabular}

}

  \end{adjustbox}
  \caption{Performance of $M-EW-L$}
\end{table*}
\begin{table*}
  \centering
  \begin{adjustbox}{width=1\textwidth, height=0.5\textheight}
  
{
\sffamily

\begin{tabular}{|c|c|c|c|c|c|c|c|c|c|c|}
	\hline
	instance\_number&20&59&137&293&605&1230&2479&4977&9972&19963\\
	\hline
	macro\_F1Measure&0.018540654&0.035612766&0.049348764&0.075559154&0.110677816&0.14785485&0.20597911&0.25531068&0.33313245&0.38833365\\
	\hline
	macro\_Precision&0.0590399&0.114397734&0.18723457&0.26407558&0.301078&0.35538006&0.3707783&0.42841172&0.51068634&0.543988\\
	\hline
	macro\_Recall&0.04441729&0.03923494&0.049584012&0.06926939&0.09159223&0.11896159&0.16423345&0.20612706&0.27215406&0.3269379\\
	\hline
	micro\_F1Measure&0.10098176&0.27577138&0.3083801&0.3467742&0.38253856&0.41619918&0.4479313&0.47335204&0.5035063&0.5440042\\
	\hline
	micro\_Precision&0.100981764&0.27577138&0.3083801&0.3467742&0.38253856&0.41619915&0.4479313&0.47335204&0.5035063&0.5440042\\
	\hline
	micro\_Recall&0.100981764&0.27577138&0.3083801&0.3467742&0.38253856&0.41619915&0.4479313&0.47335204&0.5035063&0.5440042\\
	\hline
	no\_relation\_F1Measure&0.1507579&0.42401573&0.465285&0.5024424&0.5241783&0.54419035&0.55745906&0.5710204&0.5827139&0.600622\\
\hline
\end{tabular}

}

  \end{adjustbox}
  \caption{Performance of $M-EW-LUE$}
\end{table*}
\begin{table*}
  \centering
  \begin{adjustbox}{width=1\textwidth, height=0.5\textheight}
  
{
\sffamily

\begin{tabular}{|c|c|c|c|c|c|c|c|c|c|c|}
	\hline
	instance\_number&20&59&137&293&605&1230&2479&4977&9972&19963\\
	\hline
	macro\_F1Measure&0.022839328&0.035612766&0.04638321&0.0713174&0.10786148&0.14787763&0.2020602&0.26211852&0.3279714&0.40700668\\
	\hline
	macro\_Precision&0.059927072&0.114397734&0.16126151&0.19998644&0.27103445&0.34230348&0.38716424&0.43652856&0.4971533&0.56359005\\
	\hline
	macro\_Recall&0.050523154&0.03923494&0.049035322&0.06678242&0.090275615&0.12085392&0.16093616&0.21077916&0.26671925&0.34504116\\
	\hline
	micro\_F1Measure&0.11132539&0.27577138&0.318899&0.35518935&0.37640256&0.4128682&0.4456522&0.47791025&0.5105189&0.5403226\\
	\hline
	micro\_Precision&0.11132538&0.27577138&&0.35518935&0.37640253&0.41286817&0.4456522&0.47791025&0.5105189&0.5403226\\
	\hline
	micro\_Recall&0.11132538&0.27577138&0.480424&0.35518935&0.37640253&0.41286817&0.4456522&0.47791025&0.5105189&0.5403226\\
	\hline
	no\_relation\_F1Measure&0.1849711&0.42401573&0.34880835&0.51078284&0.5227273&0.5433462&0.5615782&0.57435477&0.58848524&0.5982222\\
\hline
\end{tabular}

}

  \end{adjustbox}
  \caption{Performance of $M-EW-LUEW$}
\end{table*}



\backmatter
\bibliographystyle{anuthesis}
\bibliography{thesis}

\printindex

\end{document}